\documentclass{article} 
\usepackage{iclr2025_conference, times}

\usepackage{amsmath,amsfonts,bm}

\def\eqref#1{equation~\ref{#1}}

\def\1{\bm{1}}

\DeclareMathAlphabet{\mathsfit}{\encodingdefault}{\sfdefault}{m}{sl}
\SetMathAlphabet{\mathsfit}{bold}{\encodingdefault}{\sfdefault}{bx}{n}

\usepackage{hyperref}
\usepackage{url}
\usepackage[utf8]{inputenc} 
\usepackage[T1]{fontenc}    
\usepackage{booktabs}       
\usepackage{amsfonts}       
\usepackage{nicefrac}       
\usepackage{microtype}      
\usepackage{xcolor}         
\usepackage{subcaption}
\usepackage[hypcap=true]{caption}
\usepackage{hypcap}
\usepackage{tikz}
\usepackage{pgfplots}
\pgfplotsset{compat=1.18}

\usepackage{graphicx}
\usepackage{multirow}
\usepackage{multicol}
\usepackage{float}
\usepackage{comment}
\usepackage{amsmath}
\usepackage{amssymb}
\usepackage{mathtools}
\usepackage{amsthm}
\usepackage{bbm}
\usepackage{dsfont}
\usepackage{wrapfig}

\usepackage{xparse}
\usepackage[normalem]{ulem}

\usepackage{amssymb}

\NewDocumentCommand\NewIndexedVar{mmm}{
\expandafter\NewDocumentCommand\csname#1\endcsname{se{^_}}{#2\IfNoValueTF{##2}{}{^{##2}}\IfNoValueTF{##3}{\IfBooleanTF{##1}{}{_#3}}{_{##3}}}
}

\NewDocumentCommand\Tset{}{\mathcal{T}}

\NewDocumentCommand\goal{}{\boldsymbol{g}}

\NewDocumentCommand\p{mm}{#1\left(#2\right)}

\newcommand*\diff{\mathop{}\!\mathrm{d}}

\makeatletter
\newcommand*{\transpose}{%
  {\mathpalette\@transpose{}}%
}
\newcommand*{\@transpose}[2]{%
  \raisebox{\depth}{$\m@th#1\intercal$}%
}

\NewDocumentCommand\act{s o}{\boldsymbol{\IfBooleanTF{#1}{a}{\bar{a}}}\IfNoValueTF{#2}{}{_{#2}}}
\NewDocumentCommand\state{s o}{\boldsymbol{\IfBooleanTF{#1}{s}{\bar{s}}}\IfNoValueTF{#2}{}{_{#2}}}

\usepackage[acronym]{glossaries}
\newacronym{lfp}{LfP}{Learning from Play}
\newacronym{mtil}{MTIL}{Multi-Task Imitation Learning}
\newacronym{mlp}{MLP}{Multi-Layer Perceptron}
\newacronym{mode}{MoDE}{\textbf{M}ixture-of-\textbf{D}enoising \textbf{E}xperts Policy}
\newacronym{moe}{MoE}{Mixture-of-Experts}
\newacronym{llm}{LLM}{Large-Language-Models}
\newacronym{vlm}{VLM}{Vision-Language-Models}
\newacronym{cnn}{CNN}{Convolutional neural network}
\newacronym{gcil}{IL}{Imitation Learning}
\newacronym{sde}{SDE}{Stochastic Differential Equation}
\newacronym{ode}{ODE}{Ordinary Differential Equation}

\title{Efficient Diffusion Transformer Policies with Mixture
of Expert Denoisers for Multitask Learning}

\author{
Moritz Reuss$^{1, *}$, Jyothish Pari$^{2, *}$, Pulkit Agrawal$^{2}$, Rudolf Lioutikov$^{1}$ \\
$^{1}$Intuitive Robots Lab (IRL), Karlsruhe Institute of Technology, Germany \\
$^{2}$Department of Electrical Engineering and Computer Science (EECS), MIT, USA
}

\footnotetext[1]{Correspondence to: moritz.reuss@kit.edu}
\footnotetext[2]{Both authors contributed equally.}

\iclrfinalcopy 
\begin{document}

\maketitle

\begin{abstract}
Diffusion Policies have become widely used in Imitation Learning, offering several appealing properties, such as generating multimodal and discontinuous behavior.
As models are becoming larger to capture more complex capabilities, their computational demands increase, as shown by recent scaling laws. 
Therefore, continuing with the current architectures will present a computational roadblock. 
To address this gap, we propose Mixture-of-Denoising Experts (MoDE) as a novel policy for Imitation Learning.
MoDE surpasses current state-of-the-art Transformer-based Diffusion Policies while enabling parameter-efficient scaling through sparse experts and noise-conditioned routing, reducing both active parameters by 40\% and inference costs by 90\% via expert caching.
Our architecture combines this efficient scaling with noise-conditioned self-attention mechanism, enabling more effective denoising across different noise levels. 
MoDE achieves state-of-the-art performance on 134 tasks in four established imitation learning benchmarks (CALVIN and LIBERO). 
Notably, by pretraining MoDE on diverse robotics data, we achieve 4.01 on CALVIN ABC and 0.95 on LIBERO-90. 
It surpasses both CNN-based and Transformer Diffusion Policies by an average of $57\%$ across 4 benchmarks, while using 90\% fewer FLOPs and fewer active parameters compared to default Diffusion Transformer architectures. Furthermore, we conduct comprehensive ablations on MoDE's components, providing insights for designing efficient and scalable Transformer architectures for Diffusion Policies. Code and demonstrations are available at \url{https://mbreuss.github.io/MoDE_Diffusion_Policy/}.


\end{abstract}

\section{Introduction}

Diffusion models learn to reverse an iterative process that adds Gaussian noise to data samples~\citep{ho2020denoising, song2020score}. After training, they can generate new samples conditioned on goals like instructions or images. Recently, diffusion models have gained widespread adoption as policies for \gls{gcil}~\citep{octo_2023, reuss2023goal, chi2023diffusionpolicy}. 
\gls{gcil} is a powerful paradigm to train agents from expert demonstrations to learn versatile skills~\citep{Pomerleau-1989-15721, nair2017combining, pari2021surprising, fu2024mobile}.

Diffusion policies offer several appealing properties for \gls{gcil}: they can generate diverse multimodal behavior~\citep{jia2024towards}, scale with more data \citep{octo_2023}, and handle discontinuities in the action space~\citep{chi2023diffusionpolicy}. 
However, a major limitation is their high computational cost, resulting in slower training and inference speed as models become larger. 
Standard architectures contain hundreds of millions of parameters \citep{chi2023diffusionpolicy} and require many denoising steps to generate actions. 
Large encoder modules for images and text further increase the computational requirements for \gls{gcil} policies. 
This restricts their use in real-time robotics applications, particularly in scenarios with limited on-board computing resources, such as mobile robots.

To address these challenges, we explore \gls{moe} that can scale model capacity while using fewer FLOPs for training and inference. 
The core idea behind a sparse \gls{moe} is to utilize only a subset of the total model parameters during each forward pass. 
This is achieved by having multiple expert subnetworks and a routing model, that sparsely activates experts and interpolates their outputs, based on the input. 

We introduce \gls{mode}, a scalable and efficient \gls{moe} Diffusion Policy. 
\begin{figure}
    \centering
    \vspace{-5pt} 
    \includegraphics[width=1\columnwidth]{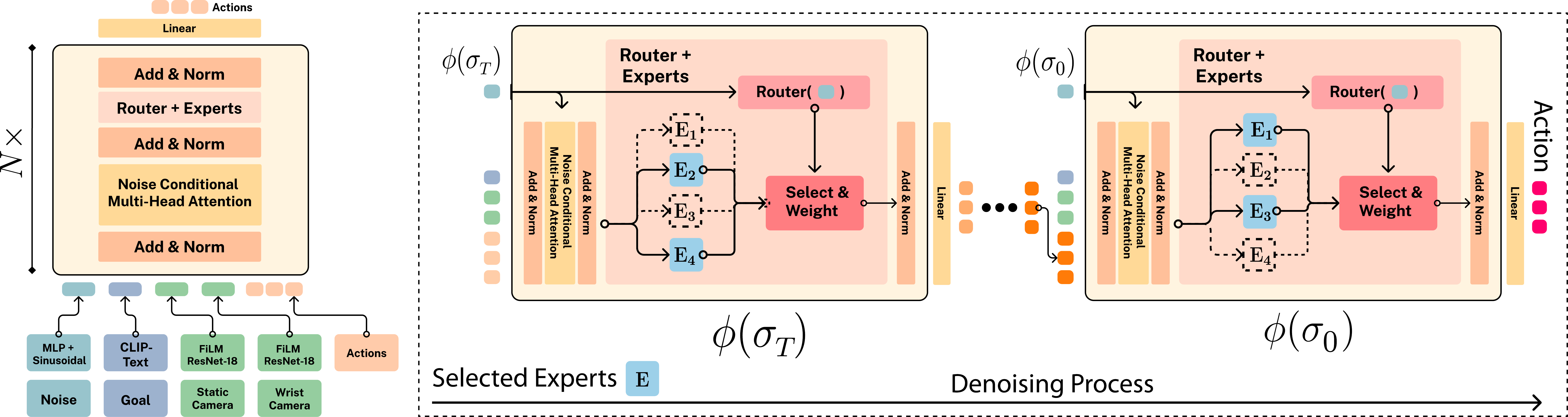}
    \caption{The proposed MoDE architecture (left) uses a transformer with causal masking, where each block includes noise-conditional self-attention and a noise-conditioned router that assigns tokens to expert models based on the noise level. This design enables efficient, scalable action generation. On the right, the router's activation of subsets of simple MLP experts with Swish-GLU activation during denoising is illustrated.}
    \label{fig: mode architecture overview}
\end{figure}
Our work is inspired by prior results showcasing the multitask nature of the denoising process~\citep{hang2024efficient}, where there is little transfer between the different phases in the denoising process. 
We present a novel noise-conditioned routing mechanism, that distributes tokens to our experts based on the current noise level.
\gls{mode} leverages noise-conditioned self-attention combined with a noise input token for enhanced noise-injection. 
Our proposed Policy surpasses previous Diffusion Policies with higher efficiency and demonstrates sota performance across 134 diverse tasks in challenging goal-conditioned imitation learning benchmarks: CALVIN~\citep{mees2022calvin} and LIBERO~\citep{liu2023libero}.
Through comprehensive ablation studies, we investigate the impact of various design decisions, including token routing strategies, noise-injection techniques, expert distribution and diverse pretraining on a large-scale robot dataset \cite{open_x_embodiment_rt_x_2023}.
We summarize our contributions below:
\begin{itemize}
    \item We introduce \gls{mode}, a novel Mixture-of-Experts Diffusion Policy that achieves state-of-the-art performance while using 90$\%$ fewer FLOPs and less active parameters than dense transformer baselines thanks to our noise-based expert caching and sparse \gls{moe} design.
    \item We demonstrate MoDE's effectiveness across 134 tasks in 4 benchmarks, showing an average $57\%$ performance increase over prior Diffusion Policies while maintaining improved computational efficiency.
    \item We present detailed ablation studies that investigate the importance of routing strategies and noise-injection, visualizing expert utilization across denoising steps to identify key components of \gls{mode}.
\end{itemize}
\section{Related Work}

\textbf{Diffusion in Robotics.}
In recent years, Diffusion Models~\citep{song2019generative, ho2020denoising, karras2022elucidating} have gained widespread adoption in the context of robotics.
They are used as a policy representation for Imitation Learning~\citep{chi2023diffusionpolicy, reuss2024multimodal, xian2023chaineddiffuser, ke20243d, li2023crossway, scheikl2023movement} and in Offline Reinforcement Learning~\citep{ajay2023is, janner2022planning, Pari-RSS-22}.
Other applications of diffusion models in robotics include robot design generation~\citep{wang2023diffusebot}, video-generation~\citep{du2023learning, Ko2023Learning, ajay2023compositional} and motion planning~\citep{carvalho2023motion, urain2023se}.
The most common architecture for using diffusion models as a policy in robotics is a \gls{cnn} with additional FiLM conditioning~\citep{perez2018film} to guide the generation based on context information. 
Recently, the transformer architecture has been adopted as a strong alternative backbone for diffusion policies, specifically in \gls{gcil}.
Examples include Octo~\citep{octo_2023}, BESO~\citep{reuss2023goal} and 3D-Diffusion-Actor \citep{ke20243d}.
Sparse-DP leverages \gls{moe} to learn specialist experts for different tasks \citep{wang2024sparse}.
However, no prior work considers using a Mixture of Experts architecture for improving the computational efficiency and inference time of the denoising process for generalist policies.


\textbf{Mixture-of-Experts.}
\gls{moe} are a class of models where information is selectively routed through the model. 
The modern version of \gls{moe} was introduced in \citep{shazeer2017outrageously}, where a routing or gating network conditionally chooses a subset of experts to send an input to. 
After Transformers \citep{vaswani2017attention} proved to be an effective model that scales well with data, they were modified to have expert feed-forward networks at each block of the model in \citep{fedus2022switch} which presented Switch Transformers. 
Switch Transformers laid the groundwork that is still widely adopted in different \gls{llm} \citep{jiang2024mixtral, du2022glam}. 
This allowed for more total parameters while keeping the forward and backward FLOPs smaller than their dense counterpart, thus yielding significant performance gains. 
However, training both the router and experts in parallel is a non-trivial optimization problem, and it can yield suboptimal solutions such as expert collapse where experts learn similar functions instead of specializing \citep{chi2022on}. 
In addition, router collapse occurs when the router selects a small subset of the experts and doesn't utilize all the experts. 
This is mitigated with load balancing losses \citep{shazeer2017outrageously, fedus2022switch} which encourage the router to distribute inputs more evenly across experts. 
Multiple works have explored different methods to perform routing, such as expert choice routing \citep{zhou2022mixtureofexperts}, differential k-selection \citep{DBLP:journals/corr/abs-2106-03760}, frozen hashing functions \citep{DBLP:journals/corr/abs-2106-04426}, and linear assignment \citep{lewis2021base}. 

\textbf{Multi Task Learning in Diffusion Models.}
It has been shown that the denoising process is multi-task \cite{hang2024efficient}. Leveraging this idea, works have adopted architectures suited for multi-task learning. Some works have explicitly scheduled which parameters are specialized to which stage in the denoising process \citep{park2023denoising, go2023towards}. In extension to this \citep{park2024switch} uses the scheduling as guidance during training but also learns how to modulate representations based on the denoising stage. Finally, some works have employed different architectures based on the denoising stage \citep{lee2024multi}.

\section{Method}

In this section, we introduce \gls{mode}, our novel MoE Diffusion Policy. 
First, we formulate the problem of learning a policy for \gls{gcil}.
Next, we summarize the framework used in Diffusion Policies and then introduce our \gls{mode} architecture with our novel noise-conditioned routing and noise-conditioned self-attention that enable efficient policy design. Finally, we explain our expert caching mechanism for efficient inference and explain the pretraining of \gls{mode}.

\subsection{Problem Formulation}

We consider the problem of learning a language-conditioned policy $\pi_{\theta}(\act | \state, \goal)$ given a dataset of robot demonstrations $\Tset$.
The policy predicts a sequence of future actions $\act = (\act*,\ldots,\act*[i+j-1])$ of length $j$, conditioned on the history of state embeddings $\state = (\state*[i-h +1],\ldots, \state*[i])$ of length $h$ and a desired goal $\goal$. 
The dataset contains $\tau \in \Tset$ trajectories, where trajectory consists of a sequence of triplets of state, actions, and goal $(\state[i], \act*[i], \goal)$.
$\goal$ is a language instruction. 
Our policy is trained to maximize the log-likelihood of the action sequence given the context of state history and goal:
\begin{align}
    \mathcal{L}_{\text{IL}} = \mathbb{E} \left[ \sum_{(\state,\act, \goal) \in \Tset} \log \p{\pi_\theta}{\act | \state, \goal}  \right].
\end{align}

\subsection{Diffusion Policy}

\gls{mode} uses the continuous-time diffusion model of EDM~\citep{karras2022elucidating} as a policy representation.
Diffusion models are a type of generative model for generating data by initially adding noise through Gaussian perturbations and then reversing this process. 
\gls{mode} applies the score-based diffusion model to represent the policy $\pi_{\theta}(\act | \state, \goal)$.
The perturbation and inverse process can be described using a stochastic differential equation:
\begin{equation}
\label{eq: conditional Karras Song SDE}
\begin{split}
\diff \act  =  \big( \beta_t \sigma_t - \dot{\sigma}_t  \big) \sigma_t \nabla_a \log p_t(\act | \state, \goal) dt + \sqrt{2 \beta_t} \sigma_t d\omega_t,
 \end{split}
\end{equation}
where $\beta_t$ controls the noise injection, $d\omega_t$ refers to infinitesimal Gaussian noise, and $p_t(\act | \state, \goal)$ is the score function of the diffusion process, that moves samples away from regions of high-data density in the forward process.
To generate new samples from noise a neural network is trained to approximate the score function $\nabla_{\act} \log p_t(\act |\state, \goal)$ via Score matching (SM)~\citep{6795935}
\begin{equation}
\label{eq: SM loss}
    \mathcal{L}_{\text{SM}} = \mathbb{E}_{\mathbf{\sigma}, \act, \boldsymbol{\epsilon}} \big[ \alpha (\sigma_t) \newline  \| D_{\theta}(\act + \boldsymbol{\epsilon}, \state, \goal, \sigma_t)  - \act  \|_2^2 \big],
\end{equation}
where $D_{\theta}(\act + \boldsymbol{\epsilon}, \state, \goal, \sigma_t)$ is the trainable neural network.
During training, we sample noise from a training distribution and add it to an action sequence.
The network predicts the denoised actions and computes the SM loss. 

After training, we can generate a new action sequence starting from random noise by approximating the reverse SDE or related ODE in discrete steps using numerical ODE integrators. 
Therefore, we sample noise from the prior $\act*[T] \sim \mathcal{N}(\mathbf{0}, \sigma_T^2 \mathbf{I})$ and iteratively denoise it.
\gls{mode} uses the DDIM-solver, a numerical ODE-solver designed for diffusion models~\citep{song2021denoising}, that allows fast denoising of actions in a few steps.
\gls{mode} uses $10$ denoising steps in all our experiments.

\subsection{Mixture-of-Experts Denoising} 

We now introduce \gls{mode}, a novel approach that employs noise-conditioned expert routing to enhance diffusion-based policies.
This novel routing mechanism enables us to precompute and fuse the required experts for more efficient inference. 
An overview of MoDE's architecture with the routing mechanism is shown in \autoref{fig: mode architecture overview}.

For language conditioning, MoDE leverages a frozen CLIP language encoder to generate a latent goal vector, while image encoding utilizes FiLM-conditioned ResNets-18/50. The model processes a sequence of input tokens $\bf{X} \in \mathbb{R}^{\text{tokens} \times D}$ and noise level $\sigma_t$. A linear projection layer $\phi(\sigma_t)$ encodes the noise level into a token, which is incorporated into $\bf{X}$. The complete MoDE architecture, $\text{MoDE}(\bf{X}, \phi(\sigma_t))$, consists of $L$ transformer blocks, each specialized for different denoising phases.

We now define each block $f^i$ as a composition of a self-attention (SA) layer and an $\text{MoE}$ layer,
\begin{equation}
    f^i(\bf{X}, \phi(\sigma_t)) = \text{MoE}(\text{SA}(\hat{\bf{X}}) + \bf{X}, \phi(\sigma_t)) + \bf{X}.    
\end{equation}

A key change in our approach is the integration of noise-aware positional embeddings. 
By adding $\phi(\sigma_t)$ to all tokens in $\bf{X}$ before self-attention:
\begin{equation}
\hat{\bf{X}} = \phi(\sigma_t) + \bf{X},
\end{equation}
we enable each token to adapt its attention patterns based on the current denoising phase. This design enhances denoising performance without introducing additional parameters or architectural complexity.

The self-attention mechanism follows the standard formulation \citep{vaswani2017attention}:
\begin{equation}
\text{SA}(\hat{\bf{X}}) = \text{softmax}(\frac{1}{\sqrt{D}} [\hat{\bf{X}}W_Q][\hat{\bf{X}}W_K]^T)[\hat{\bf{X}}W_V].
\end{equation}
Our MoE layer introduces a novel noise-conditioned routing strategy. Given $N$ experts ${\bf{E_i}}{i=1}^N$, the layer output is:
\begin{equation}
\text{MoE}(\bf{X}, \phi(\sigma_t))= \sum{i=1}^N \bf{R}(\phi(\sigma_t)) \bf{E}_i(\bf{X}),
\end{equation}
where the routing function $\bf{R}(\cdot): \mathbb{R}^{\text{tokens} \times D} \rightarrow \mathbb{R}^{\text{tokens} \times N}$ determines expert selection:
\begin{equation}
\bf{R}(\phi(\sigma_t)) = \text{topk}(\text{softmax}(\phi(\sigma_t)W_R), \textit{k})
\end{equation}
Unlike traditional MoE approaches that route based on input content, MoDE's routing mechanism specifically considers the noise level. This enables specialized experts for different denoising phases, allowing for both improved performance and computational efficiency through expert caching (detailed in \autoref{sec:router_expert_caching}). We utilize the same method from \citep{muennighoff2024olmoe} to initialize the router, which is from a truncated normal distribution with $\text{std} = 0.02$. 
We implement $\text{topk}$ using multinomial sampling without replacement, selecting $\textit{k}$ elements according to $\text{softmax}(\phi(\sigma_t)W_R)$ probabilities. To maintain gradient flow through the non-differentiable sampling process, we scale expert outputs by routing probabilities and renormalize the selected probabilities.
To prevent expert collapse, we incorporate a load balancing loss ($LB$) \citep{fedus2022switch}:
\begin{equation}
LB(\sigma_t) = N \sum_{n=1}^N \frac{1}{|\mathcal{B}|}(\sum_{i=1}^{|\mathcal{B}|}\mathbbm{1}{\bf{R}(\phi(\sigma_{t_i}))n > 0})\frac{1}{|\mathcal{B}|}(\sum{i=1}^{|\mathcal{B}|}\text{softmax}(\phi(\sigma_{t_i})W_R)_n)
\end{equation}
with $\gamma=0.01$ as the balancing coefficient.

\subsubsection{Router and Expert Caching}
\label{sec:router_expert_caching}
\begin{figure*}[t]
\hspace{-1em}
    \centering
    \includegraphics[width=0.85\columnwidth]{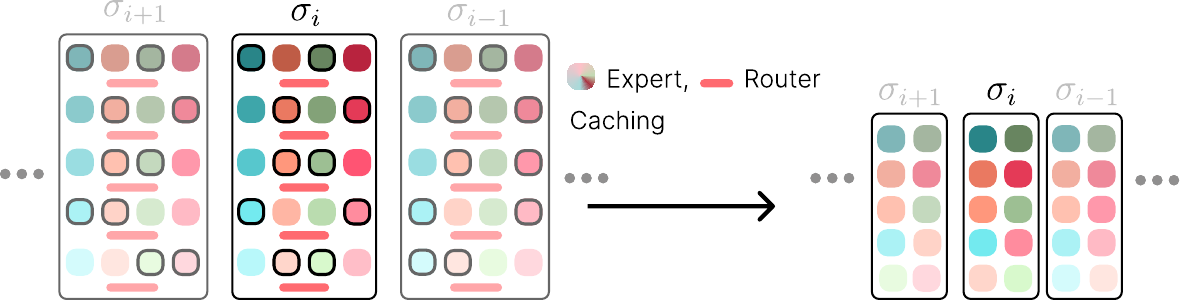}
    \caption{After training MoDE, the router is noise-conditioned, allowing pre-computation of the experts used at each noise level before inference. This enables removing the router and retaining only the selected experts, significantly improving network efficiency.}
    \label{fig:caching}
\end{figure*}
To make our method more efficient, we exploit the fact that our \gls{moe} is noise-conditioned, meaning where at each noise level, the routing path is deterministic and can be precomputed. By doing so, we can determine the chosen experts ahead of time for each noise level. \autoref{fig:caching} illustrates this process. This allows us to fuse the selected expert MLPs into a single, composite MLP, effectively reducing the computation cost.
Instead of looping through each expert individually, this fused expert MLP can run in parallel, which substantially decreases the overall latency of the network. By eliminating the need for dynamically invoking each expert, we not only save time but also streamline memory access patterns, reducing the overhead typically associated with routing decisions in a traditional \gls{moe} setup.
Our caching strategy reduces the FLOPs overhead by over 80$\%$ compared to standard \gls{moe} rollouts and makes it two times faster during inference.

\subsection{Generalist Policy Pre-Training}
\label{sec:generalist-pretraining}
We pretrain \gls{mode} on a diverse mix of multi-robot datasets curated from the OXE dataset \cite{open_x_embodiment_rt_x_2023}. 
Our training data encompasses 196k trajectories selected from six diverse datasets, featuring various robot platforms and manipulation tasks.
The pretraining process of \gls{mode} runs for 300k steps on a distributed Cluster of 6 GPUs over three days
For finetuning we freeze the pretrained routers in each layer and only finetune the other model components. 
A comprehensive overview of our pretraining dataset composition and methodology is provided in the Appendix (\autoref{sec:pretraining-details}).

In detailed evaluations using the real2sim benchmark SIMPLER\citep{li24simpler}, \gls{mode} demonstrates superior performance compared to state-of-the-art generalist policies.
It achieves an average success rate of 26.30\% across diverse manipulation tasks, outperforming both OpenVLA (23.70\%) and Octo (17.75\%). Full evaluation details are provided in \autoref{sec:simpler-eval}.

\section{Evaluation}
\label{sec:evaluation}

Our experiments aim to answer four key questions:
(I) How does \gls{mode} compare to other policies and prior diffusion transformer architectures in terms of performance?
(II) Does large-scale pre-training of diverse robotics data boost the performance of \gls{mode}?
(III) What is \gls{mode}'s efficiency and speed compared to dense transformer baselines?
(IV) Which token routing strategy is most effective for Diffusion Policies in state-based and image-based environments?
(V) How does the model utilize different experts during the action-denoising process?

We compare \gls{mode} against prior diffusion transformer architectures \citep{chi2023diffusionpolicy}, ensuring fair comparisons by using a similar number of active parameters. 
\gls{mode} uses 8 layers with 4 experts and a latent dimension of 1024 in all experiments. 
Our pretrained variant is slightly bigger with 12 layers and 4 experts with the same latent dimension of 1024.

We use an action chunking length of 10 and a history length of 1 for all experiments.
\gls{mode} does execute all 10 generated actions without early re-planning or temporal aggregation.
Detailed hyperparameters are provided in the Appendix ( \autoref{tab:hyperparameters}).

\subsection{Long-Horizon Multi-Task Experiments}
\begin{figure*}[ht]
    \centering
    \begin{subfigure}{0.35\textwidth}  
        \centering
        \includegraphics[width=\linewidth]{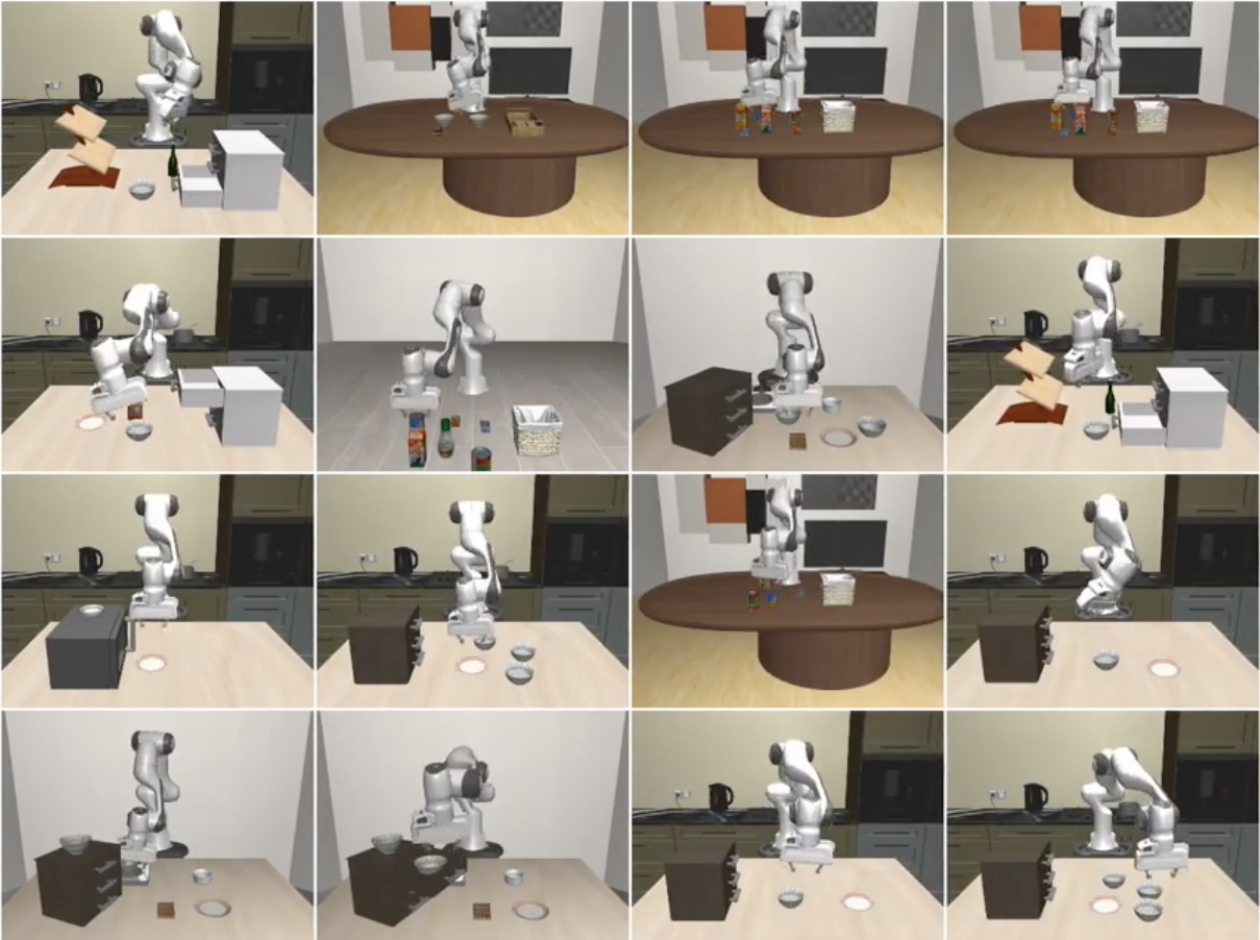}
        \caption{LIBERO-90 Tasks}
        \label{fig:libero-tasks}
    \end{subfigure}
    \hfill
    \begin{subfigure}{0.6\textwidth}  
        \centering
        \includegraphics[width=\linewidth]{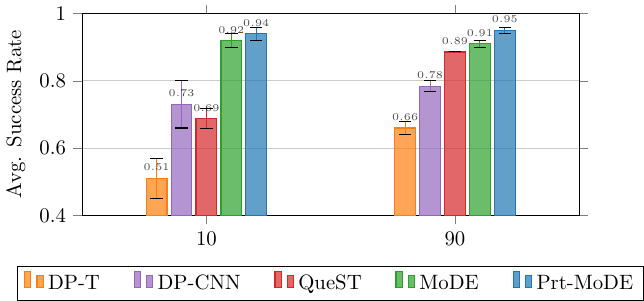}
        \caption{Results for LIBERO-10 and LIBERO-90}
        \label{fig:libero-results}
    \end{subfigure}
    \caption{Visualization and Results for LIBERO environment. (a) Few example environments and tasks of the LIBERO-90 task suite. (b) Average results for both LIBERO challenges averaged over $3$ seeds with $20$ rollouts for each task.}
    \label{fig:two-environments}
\end{figure*}

We first evaluate MoDE on the LONG-challenge and LIBERO-90 challenge of the LIBERO benchmark \citep{liu2023libero}. The LONG challenge requires a model to learn 10 tasks in different settings. 
It demands long-horizon behavior generation with several hundreds of steps for completion. The 90 variant tests policies on 90 diverse short-horizon tasks in different environments. Figure \ref{fig:libero-tasks} visualizes a few examples of these tasks.
All environments feature two cameras: a static one and a wrist-mounted camera, used to encode the current observation using FiLM-ResNets-18. 
We test each policy 20 times on each task and report the average results over 3 seeds. 
\gls{mode} and all other diffusion architectures use FiLM-conditioned ResNets-18 with a CLIP sentence embedding to encode the goal and the images.

\textbf{Baselines.}
We compare \gls{mode} against three state-of-the-art baselines:
1) The Diffusion Transformer (DP-T) architecture \citep{chi2023diffusionpolicy}, which conditions on noise and observations using a cross-attention module.
2) The standard Diffusion Policy CNN-based architecture (DP-CNN).
3) QueST \citep{mete2024quest}, a transformer-based policy that learns discrete action representations using vector-quantized embeddings of action sequences.
We tested all baselines ourselves, except for QueST, whose results are taken directly from their paper.

\textbf{Results.}
The performance of all models on the benchmark is summarized in \autoref{fig:libero-results}.
Overall, \gls{mode} achieves the highest average performance in both benchmarks, while the QueST baseline is the second best in the LIBERO-90 setting and the CNN-architecture is second best in the long horizon setting. 
These results demonstrate \gls{mode}'s ability to learn long-horizon tasks with high accuracy. 
The performance gap is more pronounced in the challenging LIBERO-10 experiment, where MoDE is the first policy to achieve an over $90\%$ success rate. 
Furthermore, \gls{mode} surpasses prior best Diffusion baselines by an average of $16\%$ in both settings, all while maintaining its computational advantage. 
The pretrained \gls{mode} variant achieves an even higher performance in both settings, showing the potential of diverse pretraining.
This showcases \gls{mode}'s ability to achieve state-of-the-art performance with a more efficient use of computational resources.

\subsection{Scaling Multi-Task Experiments}

\begin{figure*}[ht]
    \centering
    \begin{subfigure}{0.19\textwidth}  
        \centering
        \includegraphics[width=\linewidth]{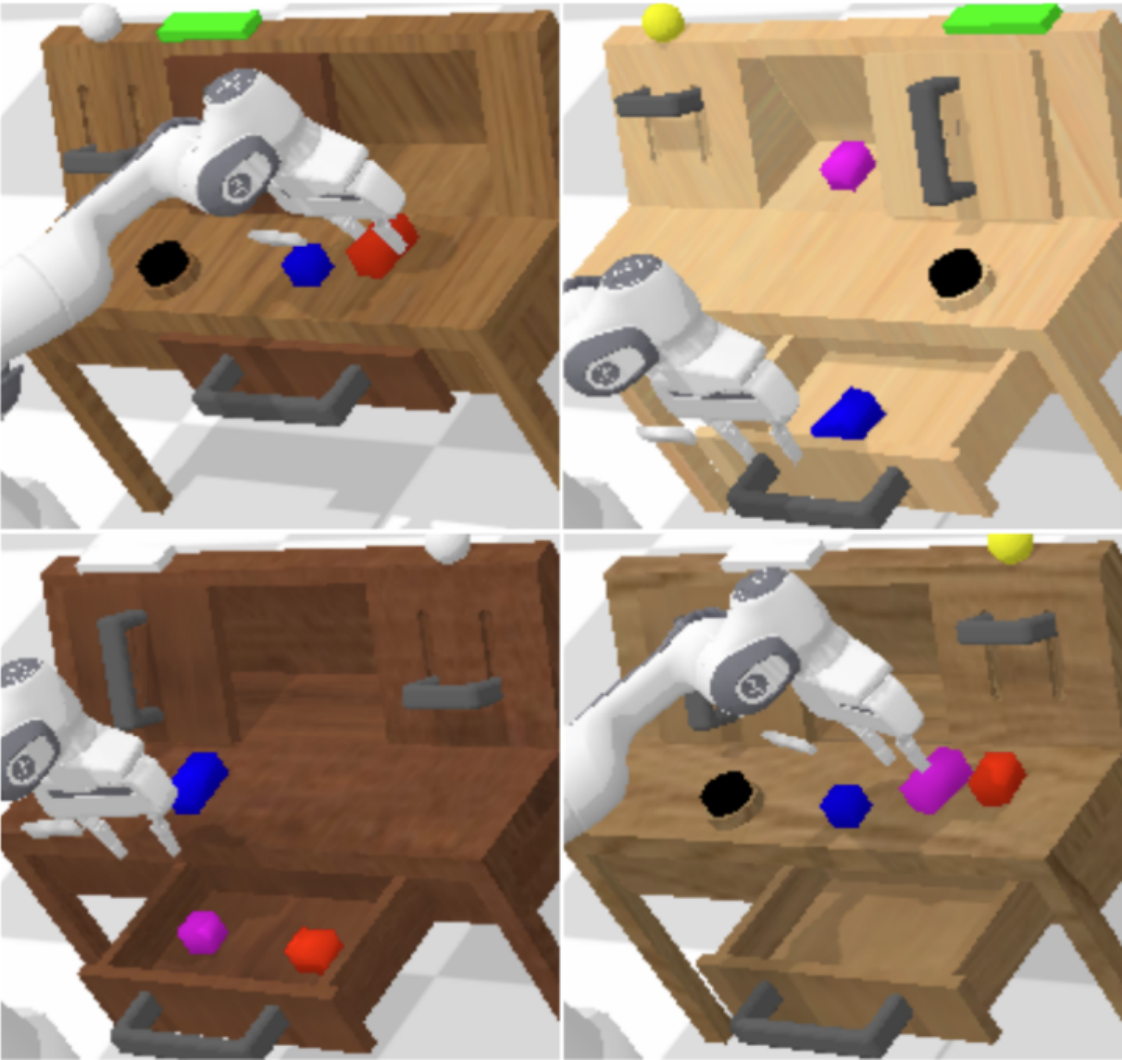}
        \caption{Environments}
        \label{fig:CALVIN-Envs}
    \end{subfigure}
    \hfill
    \begin{subfigure}{0.79\textwidth}  
        \centering
        \includegraphics[width=\linewidth]{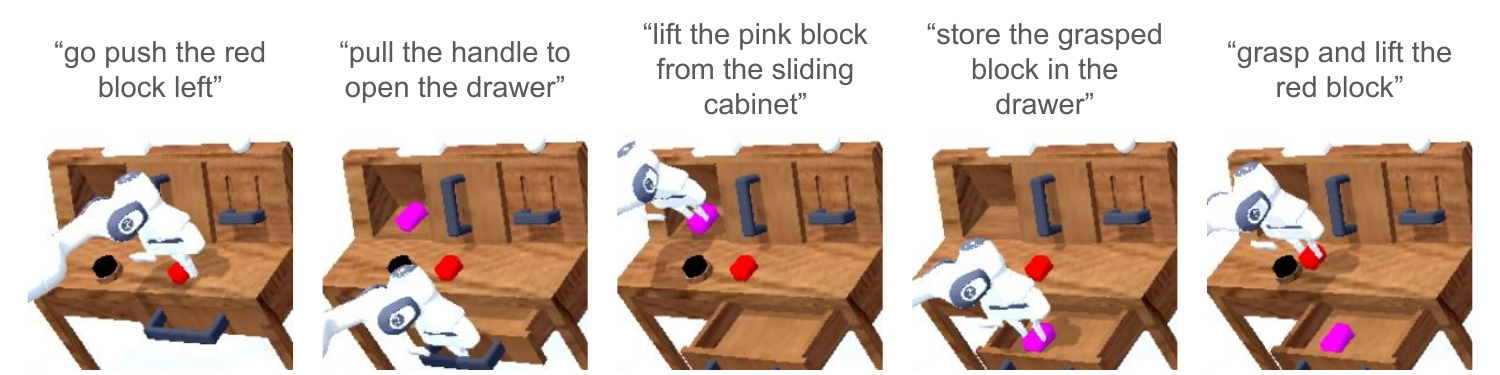}
        \caption{Example CALVIN-Rollout}
        \label{fig:CALVIN-Results}
    \end{subfigure}
    \caption{Overview of the CALVIN environment. (a) CALVIN contains four different settings (A,B,C,D) with different configurations of slides, drawers and textures. (b) Example rollout consisting of $5$ tasks in sequence. The next goal is only given to the policy, if it manages to complete the prior.}
    \label{fig:CALVIN-Overivew-Results}
\end{figure*}

\begin{table*}
\centering
\scalebox{0.8}{
\begin{tabular}{ll|cc|cccccc}
\toprule
\multirow{2}{*}{Train$\rightarrow$Test} & \multirow{2}{*}{Method} & \multirow{2}{*}{Active Params} & \multirow{2}{*}{PrT} & \multicolumn{6}{c}{No. Instructions in a Row (1000 chains)} \\
\cmidrule(lr){5-10}
 &  & in Million &  & 1 & 2 & 3 & 4 & 5 & \textbf{Avg. Len.} \\ \midrule
\multirow{6}{*}{ABCD$\rightarrow$D}
& Diff-P-CNN & 321 & $\times$ & 86.3\% & 72.7\% & 60.1\% & 51.2\% &  41.7\% & 3.16$\pm$0.06 \\
& Diff-P-T & 194  & $\times$ & 78.3\% & 53.9\% & 33.8\% & 20.4\% & 11.3\% & 1.98$\pm$0.09 \\
& RoboFlamingo & 1000 & $\checkmark$ & 96.4\% & 89.6\% & 82.4\% & 74.0\% & 66.0\% & 4.09$\pm$0.00 \\
& GR-1 & 130 & $\checkmark$ & 94.9\% & 89.6\% & 84.4\% & 78.9\% & 73.1\% & 4.21$\pm$0.00 \\
& \textbf{MoDE} & 277 & $\times$ & 96.6\% & 90.6\% & 86.6\% & 80.9\% & 75.5\% & 4.30$\pm$0.02 \\
& \textbf{MoDE} & 436 & $\checkmark$ & \textbf{97.1\%} & \textbf{92.5}\% & \textbf{87.9}\% & \textbf{83.5}\% & \textbf{77.9}\% & \textbf{4.39}$\pm$\textbf{0.04} \ \\
\midrule
\multirow{7}{*}{ABC$\rightarrow$D}
& Diff-P-CNN & 321 & $\times$ & 63.5\% & 35.3\% & 19.4\% & 10.7\% & 6.4\% & 1.35$\pm$0.05 \\
& Diff-P-T & 194 & $\times$ & 62.2\% & 30.9\% & 13.2\% & 5.0\% & 1.6\% & 1.13$\pm$0.02 \\
& RoboFlamingo & 1000 & $\checkmark$ & 82.4\% & 61.9\% & 46.6\% & 33.1\% & 23.5\% & 2.47$\pm$0.00 \\
& SuSIE & 860+ & $\checkmark$ & 87.0\% & 69.0\% & 49.0\% & 38.0\% & 26.0\% & 2.69$\pm$0.00 \\
& GR-1 & 130 & $\checkmark$ & 85.4\% & 71.2\% & 59.6\% & 49.7\% & 40.1\% & 3.06$\pm$0.00 \\
& \textbf{MoDE} & 307 & $\times$ & 91.5\% & 79.2\% & 67.3\% & 55.8\% & 45.3\% & 3.39$\pm$0.03 \\
& \textbf{MoDE} & 436 & $\checkmark$ & \textbf{96.2\%} & \textbf{88.9}\% & \textbf{81.1}\% & \textbf{71.8}\% & \textbf{63.5}\% & \textbf{4.01}$\pm$\textbf{0.04} \\
\bottomrule
\end{tabular}}
\caption{
Performance comparison on the two CALVIN challenges. 
The table reports average success rates for individual tasks within instruction chains and the average rollout length (Avg. Len.) to complete 5 consecutive instructions, based on 1000 chains.  
Zero standard deviation indicates methods without reported average performance. 
"Prt" denotes models requiring policy pretraining. 
Parameter counts exclude language encoders.
}
\label{tab: CALVIN Long horizon}
\end{table*}
Next, we evaluate \gls{mode} efficacy on the demanding \textbf{CALVIN Language-Skills Benchmark}~\citep{mees2022calvin}, an established image-based benchmark for \gls{gcil}.
This benchmark contains a large dataset of human-recorded demonstrations.
First, \gls{mode} is tested on the \textbf{ABCD$\rightarrow$D} challenge, which involves $22,966$ interaction sequences across four environments (A, B, C, D), with each consisting of $64$ timesteps and $34$ diverse tasks. 
These tasks require the acquisition of complex, sequential behaviors and the ability to chain together different skills. 
\autoref{fig:CALVIN-Envs} depicts the diverse configurations of interactive elements within these environments.
This particular challenge examines the scaling abilities of policies trained on a rich variety of data and skills across multiple settings. 
All policies are tested on $1000$ instructions chains consisting of $5$ tasks in sequence in environment D following the official protocol of CALVIN~\citep{mees2022calvin}. 
One example rollout with $5$ different tasks is visualized in \autoref{fig:CALVIN-Results}.
In terms of scoring, the model receives $1$ point for completing a task and only progresses to the next task upon completion of the prior one. 
We report the average sequence length over $3$ seeds with $1000$ instruction chains each.

\textbf{Baselines.}
We test MODE against several methods specialized for learning language-conditioned behavior and against other baseline diffusion policy architectures. 
We also compare \gls{mode} against RoboFlamingo and GR-1.
RoboFlamingo is a fine-tuned Vision-Language-Action model, that contains around $3$ billion parameters and has been pre-trained on diverse internet data.
GR-1 is a generative decoder-only Transformer pretrained on large-scale video generation and then co-finetuned on CALVIN \citep{wu2023unleashing}.
If available, we report the average performance of all prior work directly from their paper, given the standard evaluation protocol in CALVIN~\citep{mees2022hulc}.

\textbf{Results.} Our findings, outlined in Table~\ref{tab: CALVIN Long horizon} reveal that \gls{mode} outperforms all other policies in terms of average success rate. 
Moreover, \gls{mode} without pretraining outperforms well-established baselines like RoboFlamingo and GR-1, which depend on extensive internet-scale pretraining.
Our larger pretrained version achieves even higher performance.
Notably, while GR-1 uses fewer active parameters (130M compared to \gls{mode}'s 277M), it operates with a history length of 10 and 15 tokens for each timestep and uses pretrained ViTs for image embeddings. 
\gls{mode} proves more computationally efficient, requiring fewer FLOPs during inference (1.53 vs 27.5 GFLOPS for GR-1) and being equally fast although its more than 6 times larger (12.2 vs 12.6 ms).
The combination of sota performance and low computational demands positions \gls{mode} as a highly practical solution for multitask settings.

\subsection{Zero-shot Generalization Experiments}

Finally, we then extend our investigation to the \textbf{ABC$\rightarrow$D} challenge in the second phase, testing \gls{mode}'s zero-shot generalization abilities.
In this experiment, models are only trained on data from the first three CALVIN environments A,B,C and tested on the unseen setting of environment D, which has different positions of relevant objects and texture of the table. 
This requires policies, that are able to generalize their learned behavior to new environment configurations and different textures, which is especially challenging.
We evaluate \gls{mode} trained from scratch and \gls{mode} pretrained on a sub-set of Open-X-Embodiment Data.
This enables us to study the zero-shot performance and pretraining effectiveness of \gls{mode}.

\textbf{Baselines.}
For this experiment, we compare MODE against the previous CALVIN baselines, with the addition of SuSIE~\citep{black2023zero}.
A hierarchical policy utilizing a finetuned image-generation model, Instruct2Pix \citep{brooks2022instructpix2pix}, to generate goal images, which guide a low-level diffusion policy.
The high-level goal generation model requires large-scale pretraining.
SuSIE's results are based on $100$ rollouts only, without standard deviation, due to the computational cost of generating subgoal images.

\textbf{Results.}
The results of this experiment are summarized in the lower part of Table~\ref{tab: CALVIN Long horizon}.
\gls{mode} outperforms all tested baselines and surpasses all other Diffusion Policy architectures by a wide margin.
Further, by pretraining \gls{mode} on diverse robotics data, we achieve a new sota performance of 4.01.
Therefore, in response to Question (I), we affirmatively conclude that Mixture-of-Experts models not only enhance scaling performance but also demonstrate strong zero-shot generalization capabilities.
In addition, we can answer Question (II) by concluding that pretraining boost performance in challenging zero-shot settings. 

\subsection{Computational Efficiency of MoDE}
\begin{figure*}[ht]
    \includegraphics[width=0.9\linewidth]{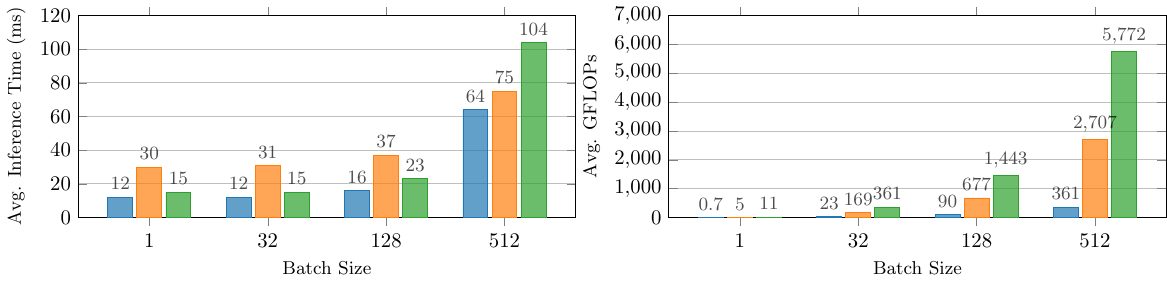}
    \vspace{-2mm}  
    \centering
    \begin{tikzpicture} 
    \definecolor{steelblue}{RGB}{31,119,180}
    \definecolor{darkorange}{RGB}{255,127,14}
    \definecolor{forestgreen}{RGB}{44,160,44}

        \begin{axis}[%
        hide axis,
        xmin=10,
        xmax=50,
        ymin=0,
        ymax=0.4,
        legend style={
            draw=white!15!black,
            legend cell align=left,
            legend columns=3, 
            legend style={
                draw=none,
                column sep=1ex,
                line width=1pt
            }
        },
        ]
        \addlegendimage{only marks, mark=square*, fill=steelblue, mark options={draw=steelblue}, mark size=3pt}
        \addlegendentry{\small MoDE w/ Cache};
        \addlegendimage{only marks, mark=square*, fill=darkorange, mark options={draw=darkorange}, mark size=3pt}
        \addlegendentry{\small MoDE w/o Cache};
        \addlegendimage{only marks, mark=square*, fill=forestgreen, mark options={draw=forestgreen}, mark size=3pt}
        \addlegendentry{\small Dense-T};
        \end{axis}
    \end{tikzpicture}
    \caption{Computational efficiency comparison between \gls{mode} and a Dense-Transformer model with the same number of parameters. 
    Left: Average inference speed over 100 forward passes for various batch sizes. 
    Right: FLOP count for \gls{mode} with router cache and without compared against a dense baseline.
    \gls{mode} demonstrates superior efficiency with lower FLOP count and faster inference thanks to its router caching and sparse expert design.}
    \label{fig:performance_comparison}
\end{figure*}
We compare \gls{mode} against a dense transformer baseline with similar parameters by measuring average inference time and FLOPs across batch sizes. As shown in Figure \autoref{fig:performance_comparison}, \gls{mode} with caching significantly improves computational efficiency - at batch size 1, inference is 20\% faster (12ms vs 15ms), while at batch size 512, MoDE requires 16x fewer FLOPs (361 vs 5,772 GFLOPS) and achieves nearly 40\% faster inference (64ms vs 104ms). 
These results demonstrate that MoDE delivers both superior task performance and substantial computational efficiency through its architecture and caching mechanism.
A detailed comparison of inference speed and FLOPS against all other baselines on CALVIN is summarized in \autoref{sec:flops-inference-speed}.

\subsection{Ablation Studies}

To thoroughly evaluate \gls{mode}'s design choices, we conducted a series of ablation studies.
These experiments address our research questions: the computational efficiency of \gls{mode} (Question III), the impact of routing strategies (Question IV), and the token distribution (Question V).

\subsubsection{What design decisions affect MoDE's performance?}

First, we assess the impact of various design decisions on \gls{mode}'s performance. 
We ablate the choice of noise-conditioning and various MoE strategies on the LIBERO-10 benchmark. 
The results are summarized in \autoref{tab:libero-ablations}.

\textbf{Noise-Injection Ablations.}
Our experiments reveal the importance of proper noise conditioning in \gls{mode}.
The full \gls{mode} model, which uses both input noise tokens and noise-conditioned self-attention, achieves the best performance with an average success rate of $0.92$.
Removing the input noise token slightly decreases performance to $0.90$, highlighting the complementary nature of both conditioning methods.
Using only the noise token for conditioning, without noise-conditioned self-attention, further reduces performance to $0.85$.
Interestingly, using FiLM conditioning \citep{perez2018film}, a common approach in image-diffusion \citep{peebles2023scalable}, yields the lowest performance in this group at $0.81$.
These results underscore the effectiveness of our proposed noise conditioning strategy in \gls{mode}, demonstrating a clear performance advantage of $0.08$ over the FiLM approach.
\begin{wraptable}{r}{0.35\linewidth}
\tiny
    \begin{tabular}{c|c}
         &  Avg. Success. \\
         \midrule
        \textbf{MoDE} & \textbf{0.92} \\
        - Input Noise Token & 0.90 \\
        - Noise-cond Attention & 0.85 \\
        FiLM Noise Conditioning & 0.81 \\
        \midrule
        TopK=1 & 0.91 \\
        Shared Expert & 0.90 \\
        $\gamma=0.1$ & 0.90 \\
        $\gamma=0.001$ & 0.86 \\
        \midrule
        256 Embed Dim & 0.86 \\
        512 Embed Dim & 0.87 \\
\bottomrule        
    \end{tabular}
    \caption{Ablation Studies for MoDE on LIBERO-10. All results are averaged over $3$ seeds with $20$ rollouts each.
    }
    \label{tab:libero-ablations}
\end{wraptable}

\textbf{\gls{moe} Ablations.}
Next, we ablate several design decisions regarding Mixture-of-Experts.
First, we test the $\text{topk}$ number of used experts.
Setting $\text{topk}$ to 1 only marginally lowers the average performance from $0.92$ to $0.91$. \gls{mode} maintains robust performance even with a single expert.
We also examine the effect of using a shared expert, where the model consistently employs the same expert in all cases.
This approach achieves a comparable average success rate of $0.90$.
Different choices for the token-distribution loss are also tested.
While \gls{mode} uses $\gamma=0.01$ as a default value, we experiment with $\gamma$ values of $0.1$ and $0.001$, which result in average success rates of $0.90$ and $0.86$, respectively.
These results indicate that a $\gamma$ value of $0.01$ strikes the best performance.

\textbf{Latent Dimension.} 
We investigate the impact of varying the latent dimension in \gls{mode}, testing dimensions of $256$, $512$, and $1024$ (our default). 
The results show that increasing the latent dimension from $256$ to $512$ yields a modest performance improvement from $0.86$ to $0.87$, while further increasing to $1024$ provides a more substantial boost to $0.92$. 
This suggests that a larger latent dimension allows \gls{mode} to capture more complex representations, leading to improved performance.

\subsubsection{Optimal Routing Strategy for Diffusion Transformers}
Next, we answer Question (III) by testing different routing strategies for our diffusion-transformer policy across several environments. 
We test two different token routing strategies:
1) Token-only conditioned Routing and
2) Noise-only Token Routing.
(1) is commonly used in LLMs, where the routing is decided based on the tokens only. 
We test these strategies in five experiments and report the average performance over $3$ seeds: Noise-only Routing achieves an average normalized performance of 0.851, slightly outperforming Token-only Routing, which achieves 0.845.
Detailed results are summarized in \autoref{tab:routing ablation} in the Appendix.
The results demonstrate the effectiveness of our proposed routing strategy. 
While the performance difference is small, Noise-only Routing offers an additional advantage: the ability to predict all used experts based on noise levels once before roll-outs, enabling faster inference, as described in \autoref{sec:router_expert_caching}. 
This is particularly beneficial for robotics applications.

\subsubsection{How does the model distribute the tokens to different experts?}
\begin{figure*}[ht]
\hspace{-1em}
    \centering
    \includegraphics[width=0.9\columnwidth]{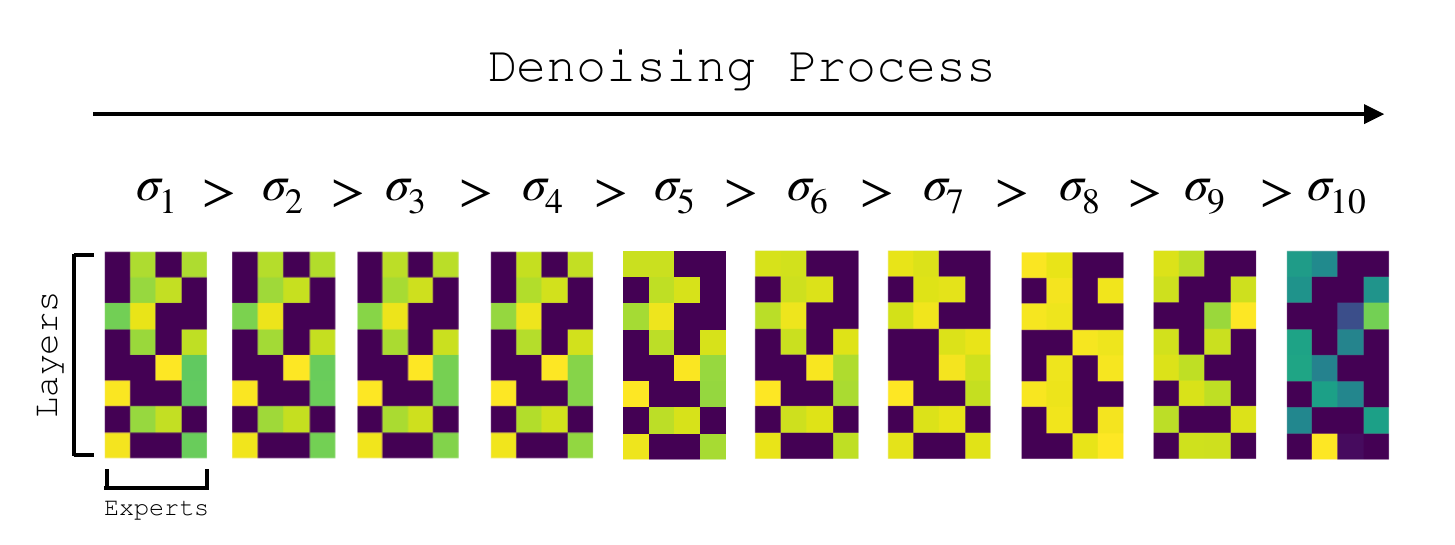}
    \caption{Visualized Expert Utilization. 
    The average usage of all experts in \gls{mode} across all layers is shown. 
    Purple color corresponds to low usage and yellow color to high one, and each image is separately normalized.  The average activation shows that \gls{mode} learns to utilize different experts for different noise levels.}
    \label{fig:expert_distribution}
\end{figure*}
To address Question IV, we analyzed how MoDE distributes tokens to different experts using a pre-trained model. 
Figure \ref{fig:expert_distribution} visualizes the average usage of each expert in each model layer during inference across various noise levels, using 10 denoising steps for clarity.
Our analysis reveals that MoDE learns to utilize different experts for various noise levels, suggesting that the router has specialized for different noise regimes. A transition in expert utilization occurs around $\sigma_8$. In the first layer, the model learns an expert specialized for low-noise levels, primarily used in the last denoising step at $\sigma_{\text{min}}$. 
We conduct more ablations studies with our pretrained model and various other versions of \gls{mode} in \autoref{sec:moe-usaage-ablations-appendix} of the Appendix.
These findings affirmatively answer Question IV, demonstrating that MoDE effectively distributes tokens across experts based on noise levels.

\subsubsection{How does the model scale with more experts?}

Finally, we analyze the effect of increasing the number of experts in \gls{mode}. 
The results are presented in \autoref{fig:n_experts_scaling}, where we evaluate \gls{mode} on the CALVIN ABCD and CALVIN ABC benchmarks using 2, 4, 6, and 8 experts. 
For comparison, we include two dense \gls{mode} baselines: Dense-small and Dense-large. Dense-small shares the same latent dimensionality as \gls{mode}, while Dense-large is scaled up to 2024 dimensions, matching \gls{mode}'s overall parameter count.
Our analysis focuses on how scaling affects both general performance (C-ABCD) and zero-shot generalization (C-ABC).
In the ABCD environment, \gls{mode} with 4 experts achieves the best performance.
Interestingly, increasing beyond 4 experts degrades performance, possibly due to overfitting or increased routing complexity.
In the zero-shot generalization (ABC), \gls{mode} with 4 experts still performs best.
Notably, the Dense-small variant consistently underperforms across both tasks, underscoring the efficiency of the MoE architecture in utilizing parameters more effectively.
We hypothesize, that 4 experts has an ideal trade-off for the context of noise-only routing of Diffusion Policies. 
The different expert specialization patterns observed in \autoref{fig:expert_distribution} and \autoref{fig:pretrained_expert_distribution} from the Appendix show that expert specialization are based on noise regions. 
\gls{mode} with more than 4 experts does not have a performance benefit.
Overall, \gls{mode} demonstrates that it can achieve comparable or superior performance to dense transformer models while requiring fewer computational resources.

\subsection{Limitations}
\label{sec:limitations}
\gls{mode} still has certain limitations. In our experiments, we find that \gls{mode} exhibits a slightly higher standard deviation compared to the baselines. 
We hypothesize that the router's initialization has significant impact on overall optimization, requiring future work on stabilizing routing models. 
In addition, when visualizing expert utilization, in some of our experiments we noticed that only a subset of the total experts were being utilized - a phenomenon known as expert collapse ~\citep{chi2022on}. 

\section{Conclusion}
\label{sec:conclusion}
In this work, we introduced Mixture-of-Denoising Experts (MoDE), a novel Diffusion Policy that leverages a mixture of experts Transformer to enhance the performance and efficiency of diffusion policies.
We also proposed a noise-conditioned routing strategy for learning specialized experts within our model.
In our extensive experiments and ablation studies across diverse benchmarks, we demonstrated the advantages of \gls{mode} to outperform prior Diffusion Policies with a lower number of parameters and $90\%$ less FLOPS during inference. 
In future work, we want to experiment with more routing strategies, such as expert-choice routing \citep{zhou2022mixtureofexperts}. 


\section{Acknowledgments}
We would like to thank Adam Wei, Anurag Ajay, Hao-Shu Fang, Anthony Simeonov, Yilun Du for their insightful discussions and feedback. 
The work was funded by the German Research Foundation (DFG) – 448648559. 
The authors also acknowledge support by the state of Baden-Württemberg through HoreKa supercomputer funded by the Ministry of Science, Research and the
Arts Baden-Württemberg and by the German Federal Ministry of Education and Research. The research was also sponsored by the Army Research Office and was accomplished under ARO
MURI Grant Number W911NF-23-1-0277.

\bibliography{mode}
\bibliographystyle{iclr2025_conference}

\appendix

\newpage

\appendix

\section{Appendix / supplemental material}

\begin{table*}[ht]
\centering
\scriptsize
\begin{tabular}{lccccc}
\toprule
\textbf{Hyperparameter} & CALVIN ABCD & CALVIN ABC & LIBERO-10 & LIBERO-90 & Pret-MoDE \\
\hline
Number of Transformer Layers & 8 & 8 & 8 & 8 & 12 \\
Number Experts & 4 & 4 & 4 & 4 & 4 \\
Attention Heads & 8 & 8 & 8 & 8 & 8 \\
Action Chunk Size & 10 & 10 & 10 & 10 & 10 \\
History Length & 1 & 1 & 1 & 1 & 1 \\
Embedding Dimension & 1024 & 1024 & 1024 & 1024 & 1024 \\
Image Encoder & FiLM-ResNet18 & FiLM-ResNet50 & FiLM-ResNet18 & FiLM-ResNet18 & FiLM-ResNet50 \\
Goal Lang Encoder & CLIP ViT-B/32 & CLIP ViT-B/32 & CLIP ViT-B/32 & CLIP ViT-B/32 & CLIP ViT-B/32 \\
Attention Dropout & 0.3 & 0.3 & 0.3 & 0.3 & 0.3 \\
Residual Dropout & 0.1 & 0.1 & 0.1 & 0.1 & 0.1 \\
MLP Dropout & 0.1 & 0.1 & 0.1 & 0.1 & 0.1 \\
Optimizer & AdamW & AdamW & AdamW & AdamW & AdamW \\
Betas & [0.9, 0.95] & [0.9, 0.95] & [0.9, 0.95] & [0.9, 0.95] & [0.9, 0.95] \\
Learning Rate & 1e-4 & 1e-4 & 1e-4 & 1e-4 & 1e-4 \\
Transformer Weight Decay & 0.05 & 0.05 & 0.05 & 0.05 & 0.1 \\
Other weight decay & 0.05 & 0.05 & 0.05 & 0.05 & 0.1 \\
Batch Size & 512 & 512 & 512 & 512 & 512 \\
Train Steps in Thousands & 30 & 25 & 15 & 30 & 300 \\
$\sigma_{\text{max}}$ & 80 & 80 & 80 & 80 & 80 \\
$\sigma_{\text{min}}$ & 0.001 & 0.001 & 0.001 & 0.001 & 0.001 \\
$\sigma_t$ & 0.5 & 0.5 & 0.5 & 0.5 & 0.5 \\
EMA & True & True & True & True & True \\
Time steps & Exponential & Exponential & Exponential & Exponential & Exponential \\
Sampler & DDIM & DDIM & DDIM & DDIM & DDIM \\
Parameter Count (Millions) & 460 & 460 & 460 & 460 & 685 \\
\bottomrule
\end{tabular}
\caption{Summary of all the Hyperparameters for the \gls{mode} policy used in our experiments.}
\label{tab:hyperparameters}
\end{table*}

\subsection{Pretraining Details}
\label{sec:pretraining-details}

\begin{table}[ht]
\centering
\begin{tabular}{lc}
\toprule
\textbf{Dataset} & \textbf{Weight} \\
\midrule
BC-Z & 0.258768 \\
LIBERO-10 & 0.043649 \\
BRIDGE & 0.188043 \\
CMU Play-Fusion & 0.101486 \\
Google Fractal & 0.162878 \\
DOBB-E & 0.245176 \\
\midrule
\textbf{Total} & 1.000000 \\
\bottomrule
\end{tabular}
\caption{Dataset sampling weights used for training \gls{mode} on a small subset of trajectories. The total dataset consists of 196k trajectories.}
\label{tab:dataset_weights}
\end{table}
We pretrain a large variant of \gls{mode} on a subset of available datasets from the Open-X-Embodiment \cite{open_x_embodiment_rt_x_2023} to study the generalization ability of \gls{mode}.
An Overview of the used dataset is summarized in \autoref{tab:dataset_weights}. 
Our pretraining dataset comprises 196K trajectories from 6 different sources, with weighted sampling across BC-Z (0.259), LIBERO-10 (0.044), BRIDGE (0.188), CMU Play-Fusion (0.101), Google Fractal (0.163), and DOBB-E (0.245). 
This dataset includes demonstrations from diverse robot platforms including Google robots, Franka Pandas, and Hello-Stretch robots, covering a wide range of manipulation tasks.
The pretraining was conducted on 6 NVIDIA A6000 GPUs with 40GB VRAM each over 3 days, completing 300K training steps. 
We used a batch size of 1024 with a learning rate of 1e-4 and weight decay of 0.1. 
To ensure balanced dataset mixing during training, we implemented a large shuffle buffer of 400K samples. 
Each dataset was individually normalized to account for different scales and ranges across the various robot platforms. 
This diverse pretraining significantly improved MoDE's zero-shot generalization, particularly on challenging benchmarks like CALVIN ABC→D where we achieved a new state-of-the-art performance of 4.01 average rollout length. 
For reproducibility, we will release our pretrained model weights and preprocessing code to the community.

For finetuning we freeze the routers of the model and remove the load-balancing loss and train on the local domain for 10k steps on LIBERO and 15k steps for CALVIN with a batch size of 64 per GPU and 4 GPUs.

\subsection{Experiments Details}

\begin{table}[t]
\centering
\label{tab:improvement-analysis}
\begin{tabular}{lccccc}
\toprule
Benchmark & MoDE & DP-T & DP-CNN & Avg. Baseline & Improvement \\
\midrule
CALVIN ABC→D (norm.) & \textbf{0.678} & 0.226 & 0.270 & 0.248 & +151.1\% \\
CALVIN ABCD→D (norm.) & \textbf{0.860} & 0.396 & 0.632 & 0.514 & +36.1\% \\
LIBERO-90 & \textbf{0.910} & 0.690 & 0.780 & 0.735 & +16.7\% \\
LIBERO-10 & \textbf{0.920} & 0.510 & 0.730 & 0.620 & +26.0\% \\
\midrule
\multicolumn{5}{l}{Average Improvement Over Second-Best:} & \textbf{57.5\%} \\
\bottomrule
\end{tabular}
\caption{Detailed Performance Improvement Analysis. CALVIN scores are normalized by dividing by 5 to align with LIBERO scale. Improvement calculated as: (MoDE - Avg. Baseline) / Avg. Baseline × 100. Final average is the mean of improvements across all four benchmarks compared to the second best Diffusion Policy variant on each one.}
\end{table}

\textbf{Average Performance Increase.}
To quantify MoDE's advantages over existing Diffusion Policies, we compared its performance against the second-best method across all benchmarks. 
MoDE demonstrated substantial improvements, particularly in challenging transfer scenarios like CALVIN ABC→D where it outperformed the next best method by 151.1\%. 
Even on the more standardized LIBERO benchmarks, MoDE maintained significant advantages of 16.7\% to 26.0\%. 
Averaging across all tasks, MoDE achieved a 57.5\% improvement over the second-best performing method while maintaining its computational efficiency with 90\% fewer FLOPs compared to dense tranformers with similar number of parameters.

\subsubsection{MoDE Evaluation on SIMPLER}
\label{sec:simpler-eval}
\begin{figure}
    \centering
    \includegraphics[width=0.8\linewidth]{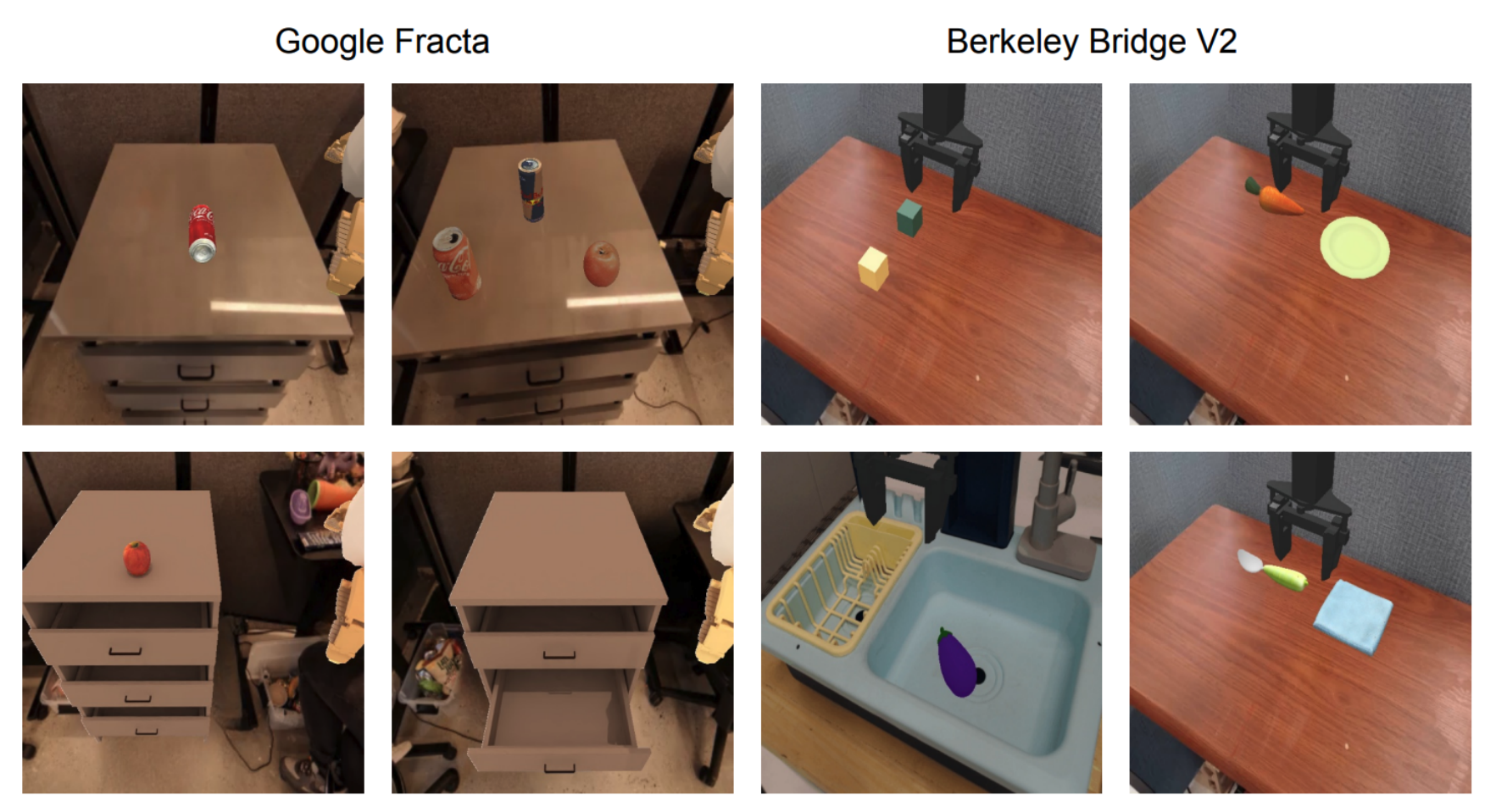}
    \caption{Example Scenes of the SIMPLER \cite{li24simpler} benchmark used to test generalist policies on various tasks from the Bridge and Google Fractal dataset.}
    \label{fig:Simpler-Envs}
\end{figure}

We evaluate MoDE's capabilities as a generalist policy by comparing it against two state-of-the-art models trained on substantially larger datasets from Open-X-Embodiment: Octo (800K trajectories) \citep{octo_2023} and OpenVLA \citep{kim2024openvla}(1M trajectories). We conduct this comparison using the SIMPLER benchmark, which provides real2sim variants of the BridgeV2 \citep{walke2023bridgedata} and Google Fractal datasets used to train RT-1 \citep{brohan2023rt}. The benchmark encompasses diverse manipulation tasks across multiple environments, as illustrated in \autoref{fig:Simpler-Envs}.

\begin{table}[h!]
\centering
\scalebox{0.9}{
\begin{tabular}{l|cc|cc|cc}
& \multicolumn{2}{c|}{\textbf{OpenVLA}} & \multicolumn{2}{c|}{\textbf{Octo Base}} & \multicolumn{2}{c}{\textbf{MoDe (ours)}} \\
\textbf{Metric} & \textbf{Score} & \textbf{Rank} & \textbf{Score} & \textbf{Rank} & \textbf{Score} & \textbf{Rank} \\
\midrule
Drawer Open & \textbf{16\%} & \textbf{1} & 0\% & 3 & 4.23\% & 2 \\
Drawer Close & 20\% & 2 & 2\% & 3 & \textbf{34.92\%} & \textbf{1} \\
Pick Can Horizontal & \textbf{71\%} & \textbf{1} & 0\% & 3 & 33.78\% & 2 \\
Pick Can Vertical & 27\% & 2 & 0\% & 3 & \textbf{29.78\%} & \textbf{1} \\
Pick Can Standing & \textbf{65\%} & \textbf{1} & 0\% & 3 & 36.44\% & 2 \\
Move Near & \textbf{48\%} & \textbf{1} & 3\% & 3 & 30\% & 2 \\
Drawer Open & 19\% & 2 & 1\% & 3 & \textbf{21.30\%} & \textbf{1} \\
Drawer Close & 52\% & 2 & 44\% & 3 & \textbf{76.85\%} & \textbf{1} \\
Pick Can Horizontal & \textbf{27\%} & \textbf{1} & 21\% & 3 & 22\% & 2 \\
Pick Can Vertical & 3\% & 3 & 21\% & 2 & \textbf{40\%} & \textbf{1} \\
Pick Can Standing & 19\% & 2 & 9\% & 3 & \textbf{35\%} & \textbf{1} \\
Move Near & \textbf{46\%} & \textbf{1} & 4\% & 3 & 45.42\% & 2 \\
Partial Put Spoon on Tablecloth & 4\% & 3 & \textbf{35\%} & \textbf{1} & 29.17\% & 2 \\
Put Spoon on Tablecloth & 0\% & 3 & 12\% & \textbf{1} & \textbf{12.5\%} & \textbf{1} \\
Partial Put Carrot on Plate & 33\% & 2 & \textbf{53\%} & \textbf{1} & 29.17\% & 3 \\
Put Carrot on Plate & 0\% & 3 & 8\% & \textbf{1} & \textbf{8.33\%} & 1 \\
Partial Stack Green Block on Yellow Block & 12\% & 2 & \textbf{32\%} & \textbf{1} & 8.33\% & 3 \\
Stack Green Block on Yellow Block & 0\% & 2 & 0\% & 2 & 0\% & 2 \\
Partial Put Eggplant in Basket & 8\% & 3 & \textbf{67\%} & \textbf{1} & 37.5\% & 2 \\
Put Eggplant in Basket & 4\% & 3 & \textbf{43\%} & \textbf{1} & 8.33\% & 2 \\
\midrule
\textbf{Average } & 23.70\% & 1.95 & 17.75\% & 2.1 & \textbf{26.30\%} & \textbf{1.65} \\
\bottomrule
\end{tabular}}
\caption{Detailed comparison of MoDE against two sota Generalist Policies OpenVLA \cite{kim2024openvla} and Octo \cite{octo_2023} tested on all SIMPLER tasks with 2952 evals.}
\label{tab:policy-comparison-simpler}
\end{table}

The results of the evaluations are summarized in \autoref{tab:policy-comparison-simpler}.
On average \gls{mode} achieves the highest average success rate of 26.30\% and the best average ranking of 1.65 across all tasks, surpassing both Octo (17.75\% success, 2.1 rank) and the 7.7B parameter OpenVLA model (23.70\% success, 1.95 rank).
MoDE shows particularly strong performance on challenging manipulation tasks like drawer manipulation (34.92\% on drawer close) and precise object interactions (40\% on vertical can picking). 
While specialized tasks like stacking blocks remain challenging for all models, MoDE's consistent performance across diverse tasks demonstrates its effectiveness as a scalable architecture for generalist policies.

\subsubsection{CALVIN Benchmark}

The CALVIN benchmark \citep{mees2022calvin} is an established \gls{gcil} benchmark for learning language-conditioned behavior from human play data.
In contrast to other benchmarks the data does not contain structured demonstrations, where the robot completes one task, but instead, the dataset was collected by humans, that randomly interact with the environment. 
From these long-horizon trajectories across the $4$ different settings, the authors randomly cut out short sequences with $64$ frames and labeled them with the task label. 
While the dataset offers models to train on the unlabeled part too, we restricted \gls{mode} to only train on the labeled parts. 
The Franke Emika Panda robot is controlled using a Delta-End-Effector Space with a discrete gripper. 
We use two cameras to encode the current scene: a static camera and a wrist one and predict the next $10$ actions, before receiving the next observations and generating another set of $10$ actions.

\textbf{CALVIN ABC.}
We train \gls{mode} and our dense transformer baseline for $25$k training steps with a batch size of $512$ on a $4$ GPU Cluster Node with $4$ A$6000$ NVIDIA GPUs for $6.5$ hours with all $1000$ rollouts at the end of training. We report the mean results averaged over $3$ seeds as done in all relevant prior work. 
All baselines are reported from the original paper given the standardized evaluation protocol of CALVIN \citep{mees2022calvin}.

\textbf{CALVIN ABCD.}
We train \gls{mode} and our dense transformer baseline for $30$k training steps with a batch size of $512$ on a $4$ GPU Cluster Node with $4$ A$6000$ NVIDIA GPUs for $7.5$ hours with all $1000$ rollouts at the end of training. We report the mean results averaged over $3$ seeds as done in all relevant prior work. 

\subsubsection{LIBERO Benchmark}

\textbf{LIBERO-10.}
The LIBERO-10 benchmark consists of $50$ demonstrations for $10$ different tasks that are all labeled with a text instruction. 
The Franka Emika Panda robot is controlled using an end-effector controller.
Similar to CALVIN all models have access to two camera inputs: a static one and a wrist camera. 
We train \gls{mode} and our dense transformer baseline for $50$  epochs with a batch size of $512$ on a $4$ GPU Cluster Node with $4$ A$6000$ NVIDIA GPUs for $2$ hours with all $200$ rollouts at the end of training.
The benchmark does require to test models on $10$ different long-horizon tasks. 
We test each task $20$ times for each model and report the final average performance overall $10$ tasks. 

\textbf{LIBERO-90.}
The LIBERO-10 benchmark consists of $50$ demonstrations for $90$ different tasks that are all labeled with a text instruction. 
The Franka Emika Panda robot is controlled using an end-effector controller.
We train \gls{mode} and our dense transformer baseline for $50k$  steps with a batch size of $512$ on a $4$ GPU Cluster Node with $4$ A$6000$ NVIDIA GPUs for $12$ hours with all $1800$ rollouts at the end of training.
The benchmark does require to test models on $90$ different tasks in many different environments. 
We test each task $20$ times for each model and report the final average performance overall $90$ tasks.

\begin{table*}
\centering
\small
\begin{tabular}{l|cc|cc|c|c}
\toprule
\textbf{Model} & \textbf{Block Push} & \textbf{Relay Kitchen} & \textbf{CAL ABC} & \textbf{CAL ABCD} & \textbf{L-10} & \textbf{Average} \\
\midrule
Dense T & 0.96$\pm$0.02 & 3.73$\pm$0.12 & \textbf{2.83}$\pm$\textbf{0.19} & 4.13$\pm$0.11 & 0.91$\pm$0.02 &  0.839$\pm$0.144 \\

Token-Router & 0.97$\pm$0.01 & \textbf{3.85}$\pm$\textbf{0.03} & 2.67$\pm$0.04 & 4.29$\pm$0.08 & 0.90$\pm$0.01 & 0.845$\pm$0.161 \\

$\sigma_t$-Router & \textit{0.97}$\pm$\textit{0.01} & 3.79$\pm$0.04 & \textit{2.79}$\pm$\textit{0.16} & \textbf{4.30}$\pm$\textbf{0.02} & \textbf{0.92}$\pm$\textbf{0.02} & \textbf{0.851}$\pm$\textbf{0.151} \\
\bottomrule
\end{tabular}
\caption{Overview of the performance of all different token routing strategies used for \gls{mode} across $5$ benchmarks.
We mark the best result for each environment in \textbf{bold} and the second best in \textit{cursive}. We use CAL to represent CALVIN.
To average the results, we normalize all scores and compute the average over all environments.}
\label{tab:routing ablation}
\end{table*}

\subsection{Baselines}

Below we explain several baselines used in the experiments in detail:

\textbf{Diffusion Policy-CNN/T}
Inspired by \citep{chi2023diffusionpolicy}, we evaluate extension of the DDPM based Diffusion Policy framework for goal-conditioned Multi-task learning. 
We evaluate two versions: the CNN-based variant and the Diffusion-Transformer variant, that is conditioned on context and noise using cross-attention. 
For our experiments we also use EDM-based Diffusion framework for fair comparison against \gls{mode}.
We optimized the ideal number of layers and latent dimension for the Transformer baseline and our final version uses $8$ layers with a latent dimension of $1024$.
Larger or smaller variants resulted in lower average performance.

\textbf{RoboFlamingo.}
RoboFlamingo \citep{li2023vision} is a \gls{vlm} finetuned for behavior generation. 
The authors use a $3$ billion parameter Flamingo model \citep{alayrac2022flamingo} and fine-tune it on CALVIN by freezing the forward blocks and only fine-tuning a new Perceiver Resampler module to extract features from a frozen vision-transformer image encoder and the cross-attention layers to process the image features.
Finally, a new action head is learned to generate actions. 
Overall, the finetuning requires training approx. $1$ billion of the parameters. 
We report the reported results from the paper since they use the standard CALVIN evaluation suite. 

\textbf{SuSIE.}
This model first finetunes Instruct2Pix, an image-generation diffusion model, that generates images conditioned on another image and a text description \citep{brooks2022instructpix2pix} on the local CALVIN robotics domain and uses it as a high-level goal generator.
The low-level policy is a \gls{cnn}-based Diffusion Policy, that predicts the next $4$ actions given the current state embedding and desired sub-goal from the high-level policy \citep{black2023zero}.

\textbf{GR-1}
A causal GPT-Transformer model \citep{wu2023unleashing}, that has been pretrained on large-scale generative video prediction of human videos. 
Later, the model is finetuned using co-training of action prediction and video prediction on CALVIN. 
We report the results directly from their paper for the CALVIN benchmark.

\subsection{Average FLOPs computation and Inference Speed}
\label{sec:flops-inference-speed}
\begin{table*}[t]
\centering
\scalebox{0.8}{
\begin{tabular}{l|ccc|c|c|c|c}
\toprule
Method & Active Params (M) & Total Params (M) & GFLOPS & PrT & Avg. Length & SF-Ratio & Inf. Time [ms] \\
\midrule
Diff-P-CNN & 321 & 321 & 1.28 & $\times$ & 1.35 & 1.05 & \textbf{11.7} \\
Diff-P-T & 194 & 194 & 2.16 & $\times$ & 1.13 & 0.53 & 16.2 \\
RoboFlamingo & 1000 & 1000 & 690 & $\checkmark$ & 2.47 & 0.004 & 65 \\
SuSIE & 860+ & 860+ & 60 & $\checkmark$ & 2.69 & 0.045 & 199 \\
GR-1 & \textbf{130} & \textbf{130} & 27.5 & $\checkmark$ & 3.06 & 0.11 & 12.6 \\
\textbf{MoDE (ours)} & 436 & 740 & 1.53 & $\checkmark$ & \textbf{4.01} & \textbf{2.6} & 12.2 \\
\bottomrule
\end{tabular}}
\caption{Comparison of total and active number of parameters of methods used in the CALVIN benchmark. Additional overview of average FLOPS required by the different methods together with their average performance on the ABC benchmark. SF-Ratio compares average rollout length with GFLOPS.}
\label{tab:CALVIN_flops_comparison}
\end{table*}

We provide an in-depth comparison of the total parameters and FLOPs used for every method in \autoref{tab:CALVIN_flops_comparison}. Additionally, we provide the computational efficiency metric per GFLOPS ($10^9$ FLOPS) to compare the various methods and measure the average prediction time for a single action.
In the following, we detail the average GFLOPS computation for all relevant baselines on the CALVIN ABC benchmark. Specifically, we compare the average GFLOPS required to predict a single action.

To guarantee a fair comparison we evaluated all methods on the same NVIDIA A6000 GPU with 40 GB VRAM. 
To compute the average inference speed, we tested each method 100 times and removed large outliers to compute an average time.

\textbf{MoDE.} 
We benchmark the large pretrained variant with 12 layers, 4 experts and a hidden dim of 1024.
The average GFLOPS for a forward pass are 0.7 GFLOPS.
Without router caching, the model would require 5 GFLOPS, indicating that the router caching reduces the overall computational cost by over 90$\%$. 
The architecture processes 14 tokens in total (1 noise + 1 goal + 2 images + 10 noisy action actions). 
\gls{mode} predicts a sequence of 10 actions with 10 denoising passes. 
For the variant with ResNet-50, the image encoder requires 8.27 GFLOPS . 
On average for a single action, \gls{mode} requires 10 forward passes with the transformer and a single pass  with the ResNet-50. 
Consequently, the pretrained variant of \gls{mode} needs 1.53 GFLOPS on average to predict a single action.
The inference time for that model depends on the hardware. 
We measure an average inference time per action of 12.2.

\textbf{DP-CNN/T.} 
DP-CNN utilizes 0.8 GFLOPS for an average forward pass. 
The ResNet-18s require 3.62 GFLOPS.
The model predicts 10 actions with 10 denoising steps and executes 10 actions without replanning. 
This results in the CNN version requiring 1.28 GFLOPS to predict a single action. 
For the Transformer version, the architecture predicts 10 actions using 10 denoising steps and processes 14 tokens in total (1 noise + 1 goal + 2 images + 10 noisy action actions) similar to \gls{mode}.  
It achieves an average GFLOPS usage of 1.8 GFLOPS per forward pass through the transformer. 
The DP-T baseline requires 2.16 GFLOPS on average to predict a single action.
The CNN version requires 11.7 ms to predict a single action on average and the transformer variant with its cross-attention conditioning needs 16.2 ms.

\textbf{RoboFlamingo.}
For computational analysis, the model requires 34 GFLOPS to encode a single image with a ViT. 
For the policy backbone, we evaluated the "mpt-1b-redpajama-200b-dolly" variant as used in the paper. 
This architecture requires 656 GFLOPS for an average sequence of 32 tokens per forward pass. 
While multiple variants of RoboFlamingo exist, this provides a rough estimate of the required GFLOPS. 
In total, we estimate an average of 690 GFLOPS to predict an action in CALVIN.
To predict a single action, the model requires 65 ms on average.

\textbf{SuSIE.}
In our computational analysis, we tested Instruct2Pix with 50 denoising steps as implemented by SuSIE. 
The resulting 1026 GFLOPS are divided by 20 as the model only generates new subgoals every 20 timesteps. 
The low-level policy uses a ResNet-50 image encoder with 8.27 GFLOPS. 
Contrast to other policies, SuSIe only computes one image per state and predicts actions every timestep. 
These are then averaged using exponential averaging.
Thus, we omit the small diffusion head to get an estimate of 60 GFLOPS per action.
For the average inference speed we meassure the time to generate a single goal image and divide it by 20, next we  add the average time to encode two images with the ResNet-50 and 10 forward passes through a small MLP.
Every 20 timesteps, when SuSIE generates a new image it requires 3777.62 ms for a single action generation.
Otherwise its a lot faster with 10.7 ms. 
On average SuSIE requires 199 ms to generate a single action, which make it the slowest policy overall.

\textbf{GR-1.}
The pretrained MAE Vision Transformer requires approx. 17.5 GFLOPS to encode a single image. 
The transformer backbone processes 150 tokens with a history length of 10 and 15 tokens per timestep (10 image tokens + 1 goal token + 1 proprioceptive token + 2 video readout tokens + 1 action token per timestep). 
Consequently, the average GFLOPS for a single action prediction utilizing the decoder with a latent dimension of 384 and 12 layers are 10 GFLOPS. 
In total, this results in an average computational cost of 27.5 for predicting a single action in CALVIN.
For the average single action prediction the model requires 12.6 ms.

\textbf{Analysis.}
Overall, \gls{mode} offers the best performance to GFLOPS ratio across all tested baselines. 
Although \gls{mode} is significantly larger in total size compared to the other Diffusion Policy architectures it requires similar inference speed and a low FLOP count. 
Additionally, it has demonstrated superior efficiency in terms of computational resources while maintaining high performance on the CALVIN benchmark tasks. 
For inference speed \gls{mode} is the second fastest although it has a high total parameter count.

\subsection{Detailed Experimental Results}

We summarize the ablations regarding the choice of routing in \autoref{tab:routing ablation}.
Therefore, we test two $2$ different routing strategies across $5$ benchmarks.

\subsection{State-based Experiments}
\noindent
We conduct additional experiments with \gls{mode} on two established multi-task state-based environments:

\textbf{Relay Kitchen.} We utilize the Franka Kitchen environment from \citep{lynch2019learning} to evaluate models. 
This virtual kitchen environment allows human participants to manipulate seven objects using a VR interface: a kettle, a microwave, a sliding door, a hinged door, a light switch, and two burners. The resulting dataset consists of $566$ demonstrations collected by the original researchers, where each participant performed four predetermined manipulation tasks per episode.
The Franka Emika Panda robot is controlled via a $9$-dimensional action space representing the robot's joint and end-effector positions. 
The $30$-dimensional observation space contains information about the current state of the relevant objects in the environment.
As a desired goal state, we randomly sample future states as a desired goal to reach. 

For this experiment, we train all models for $40$k training steps with a batch size of $1024$ and evaluate them $100$ times as done in prior work \citep{shafiullah2022behavior, cui2023from, reuss2023goal} to guarantee a fair evaluation. All reported results are averaged over $4$ seeds.
We train our models on a local PC RTX with an RTX $3070$ GPU for approx. $2$ hours for each run with the additional experimental rollouts.

\textbf{Block Push.}
The PyBullet environment features an XArm robot tasked with pushing two blocks into two square targets within a plane. 
The desired order of pushing the blocks and the specific block-target combinations are sampled from the set of $1000$ demonstrations as a desired goal state.
The demonstrations used for training our models were collected using a hard-coded controller that selects a block to push first and independently chooses a target for that block. 
After pushing the first block to a target, the controller pushes the second block to the remaining target. 
This approach results in four possible modes of behavior, with additional stochasticity arising from the various ways of pushing a block into a target. 
The models only get a credit, if the blocks have been pushed in the correct target position and order. 
We consider a block successfully pushed if its center is within 0.05 units of a target square.

All models were trained on a dataset of $1000$ controller-generated demonstrations under these randomized conditions.
All models have been trained for $60$k steps with a batch size of $1024$.
To evaluate them we follow prior work \citep{shafiullah2022behavior, cui2023from, reuss2023goal} and test them on $100$ different instructions and report the average result over $4$ seeds.
We train our models on a local PC RTX $3070$ GPU for approx. $3$ hours for each run with a final evaluation.
Demonstrations are sourced from a scripted oracle, which first pushes a randomly chosen block to a selected square, followed by the other block to a different square. 
The policies are conditioned to push the blocks in the desired configuration using a goal state-vector. 
We chose an action sequence length of $1$ given a history length of $4$ for these experiments, which are inspired by our dense diffusion transformer baseline BESO~\citep{reuss2023goal}.

\begin{wraptable}{r}{0.48\linewidth}
\begin{tabular}{l|cc}
\toprule
& Block Push & Relay Kitchen \\
\midrule
C-BeT & 0.87$\pm$(0.07) & 3.09$\pm$(0.12) \\
VQ-BeT & 0.87$\pm$(0.02) & 3.78$\pm$(0.04) \\
BESO & 0.96$\pm$(0.02)  & 3.73$\pm$(0.05) \\
\textbf{MoDE} & \textbf{0.97}$\pm$(\textbf{0.01}) & \textbf{3.79}$\pm$(\textbf{0.02}) \\
\bottomrule
\end{tabular}
\captionof{table}{Comparison of the performance of different policies on the state-based goal-conditioned relay-kitchen and block-push environment averaged over $4$ seeds. \gls{mode} outperforms the dense transformer variant BESO and other policy representations on all baselines.}
\label{tab: transposed state-based results}
\end{wraptable}
\textbf{Baselines.}
In this setting, we compare \gls{mode} against several SOTA goal-conditioned policies. 
We test two transformer architectures, C-BeT~\citep{cui2023from} and VQ-BeT~\citep{lee2024behavior}, that predict discretized actions with an offset. 
C-BeT uses k-means clustering together with an offset vector while VQ-BeT leverages residual Vector Quantization to embed actions into a hierarchical latent space. 
Further, we test against a dense diffusion policy transformer model BESO~\citep{reuss2023goal}.
BESO uses the same continuous-time diffusion policy combined with a dense transformer to predict a single action given a sequence of prior states. 
To enable a fair comparison, we chose the same hyperparameters for BESO and \gls{mode} in both settings. 
We test all models averaged over 4 seeds and report the mean values directly from prior work~\citep{lee2024behavior}. 

\textbf{Results.}
The results of both experiments are summarized in Table~\ref{tab: transposed state-based results}.
\gls{mode} achieves a new SOTA performance on both benchmarks and outperforms the dense transformer variant of BESO in both settings.
Further, \gls{mode} achieves higher performance compared to other policy representation methods such as VQ-BeT and C-BeT.

\subsubsection{Mixture-of-Experts Ablations}
\label{sec:moe-usaage-ablations-appendix}

\begin{figure}
    \centering
    \includegraphics[width=0.8\linewidth]{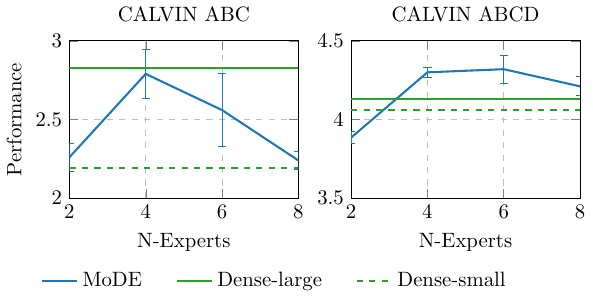}
    \caption{Scaling performance of MoDE and Dense-MoDE on CALVIN ABC and ABCD environments, showing average performance for $2$ to $8$ experts using best-performing variants for each environment.}
    \label{fig:n_experts_scaling}
\end{figure}

\textbf{Q: How does the Load Balancing Loss influence the Expert Distribution?}

\begin{figure}[htb]
    \centering
    \subfloat[$\gamma_{LB}=0.1$]{%
        \includegraphics[width=0.47\textwidth]{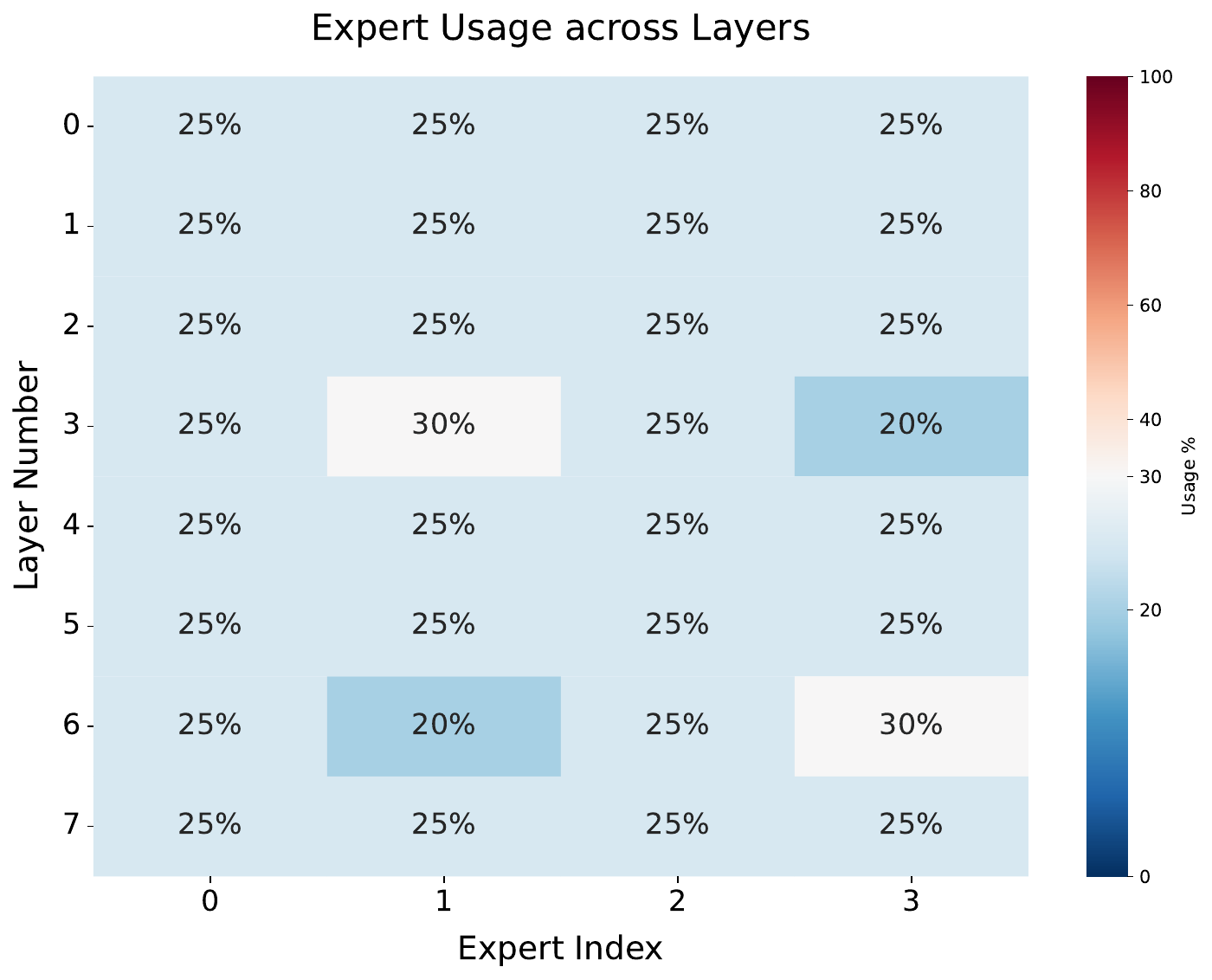}
        \label{fig:lb0.1}
    }
    \subfloat[$\gamma_{LB}=0.01$]{%
        \includegraphics[width=0.47\textwidth]{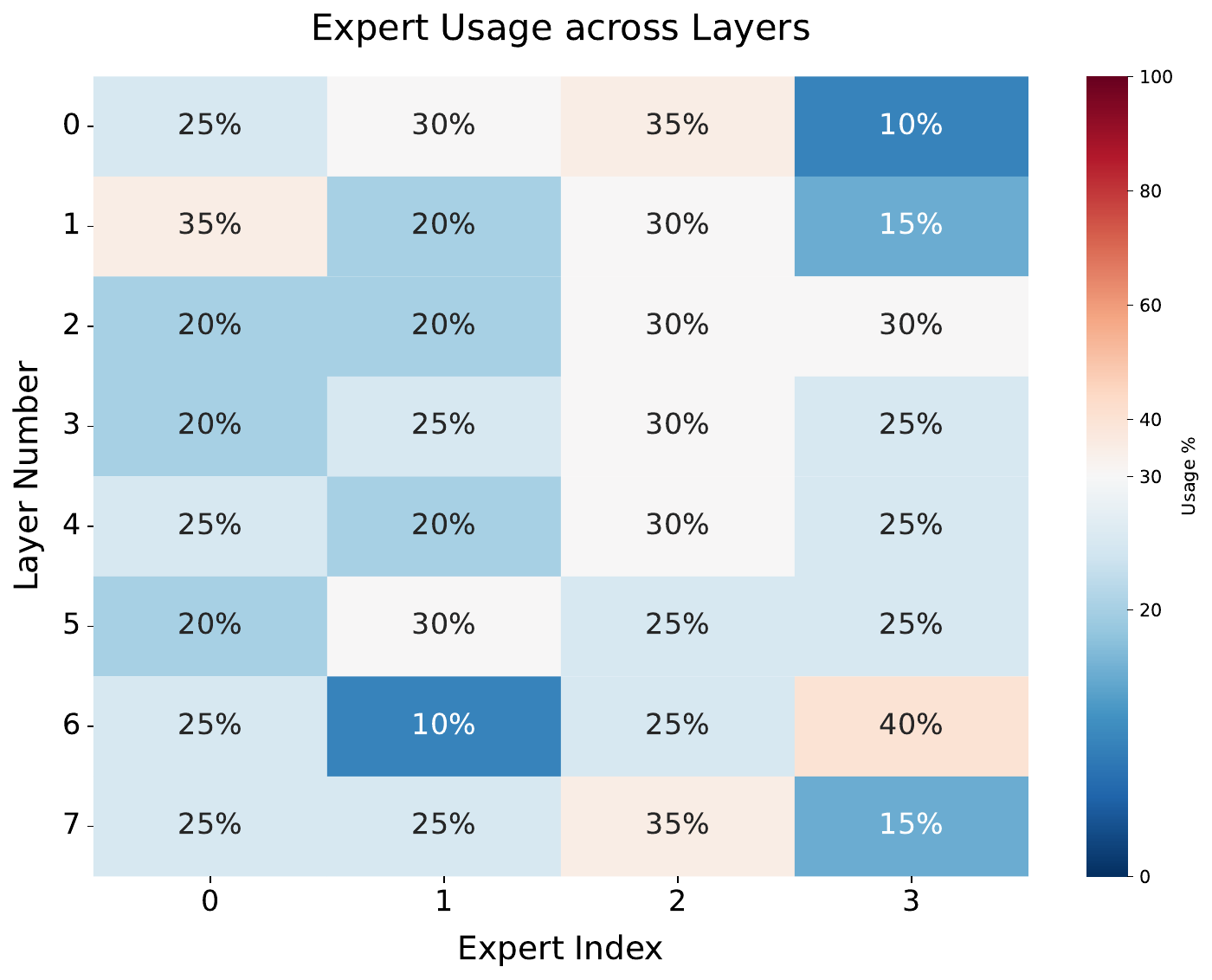}
        \label{fig:lb0.01}
    }
    \vspace{1em} 
    \subfloat[$\gamma_{LB}=0.001$]{%
        \includegraphics[width=0.47\textwidth]{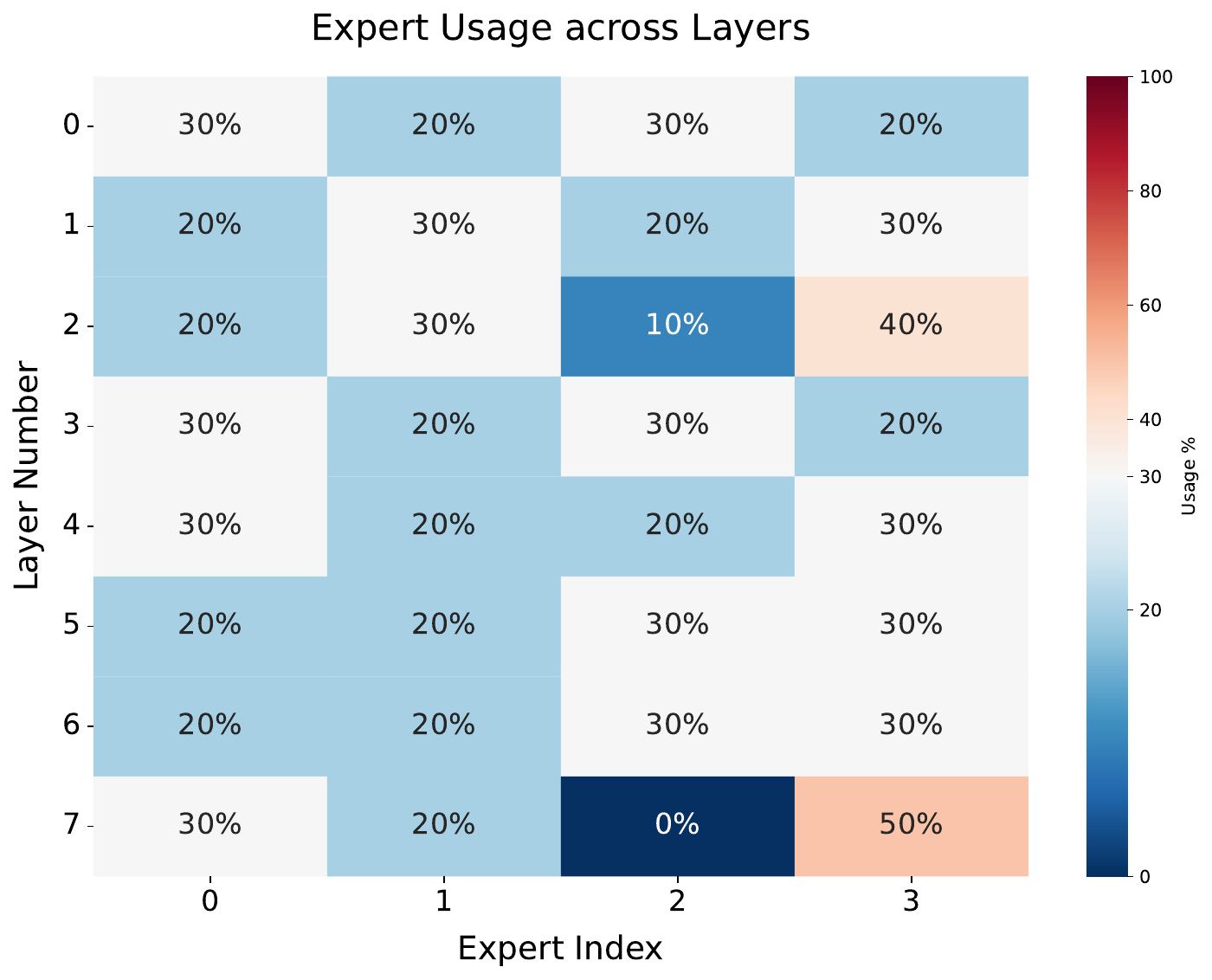}
        \label{fig:lb0.001}
    }
    \subfloat[$\gamma_{LB}=0.0001$]{%
        \includegraphics[width=0.47\textwidth]{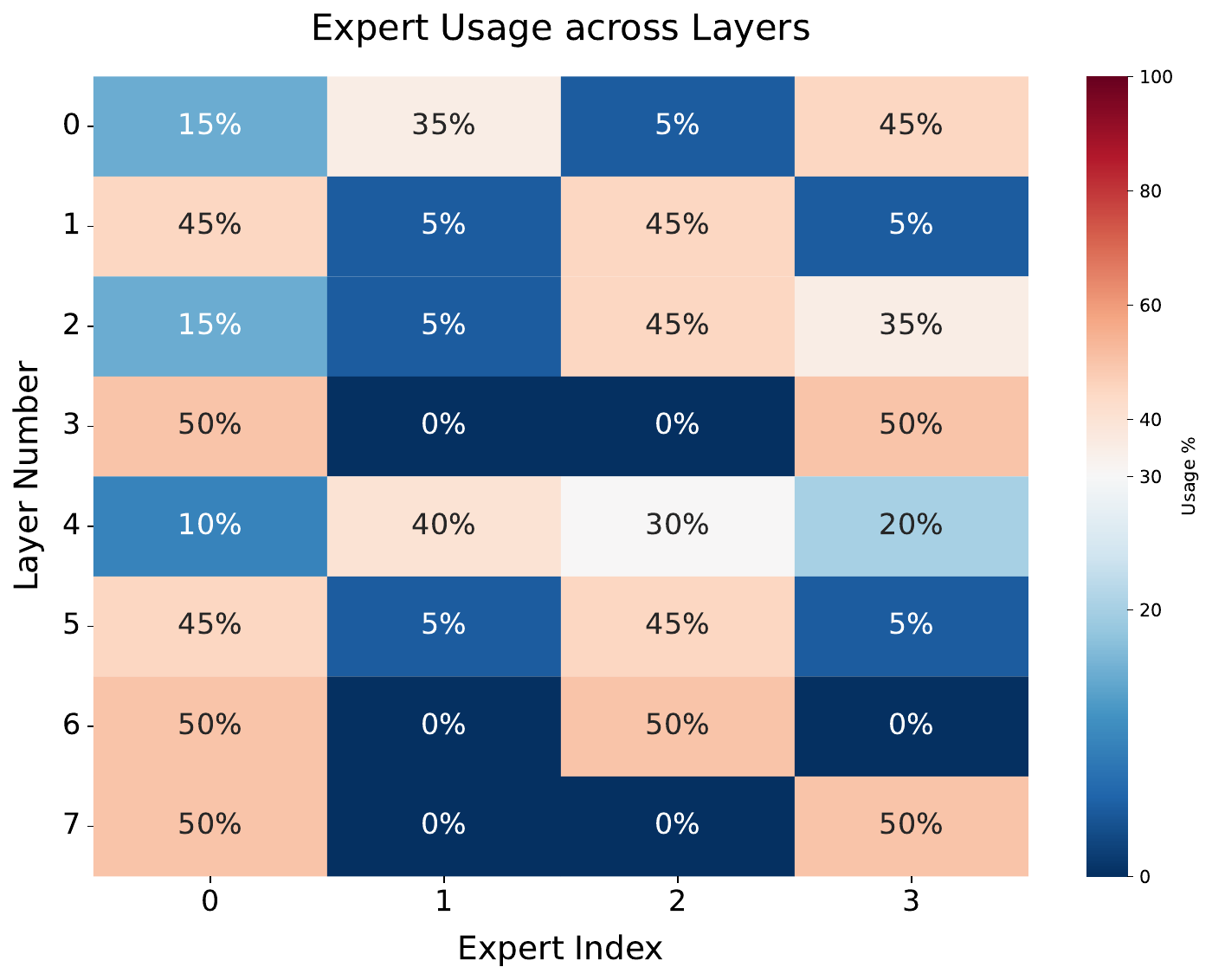}
        \label{fig:lb0.0001}
    }
    \caption{Average Expert Utilization for different Load Balancing Weights across all denoising levels.}
    \label{fig:Expert_Distribution LB}
\end{figure}

\begin{figure}[htb]
    \centering
    \subfloat[3 Experts]{%
        \includegraphics[width=0.47\textwidth]{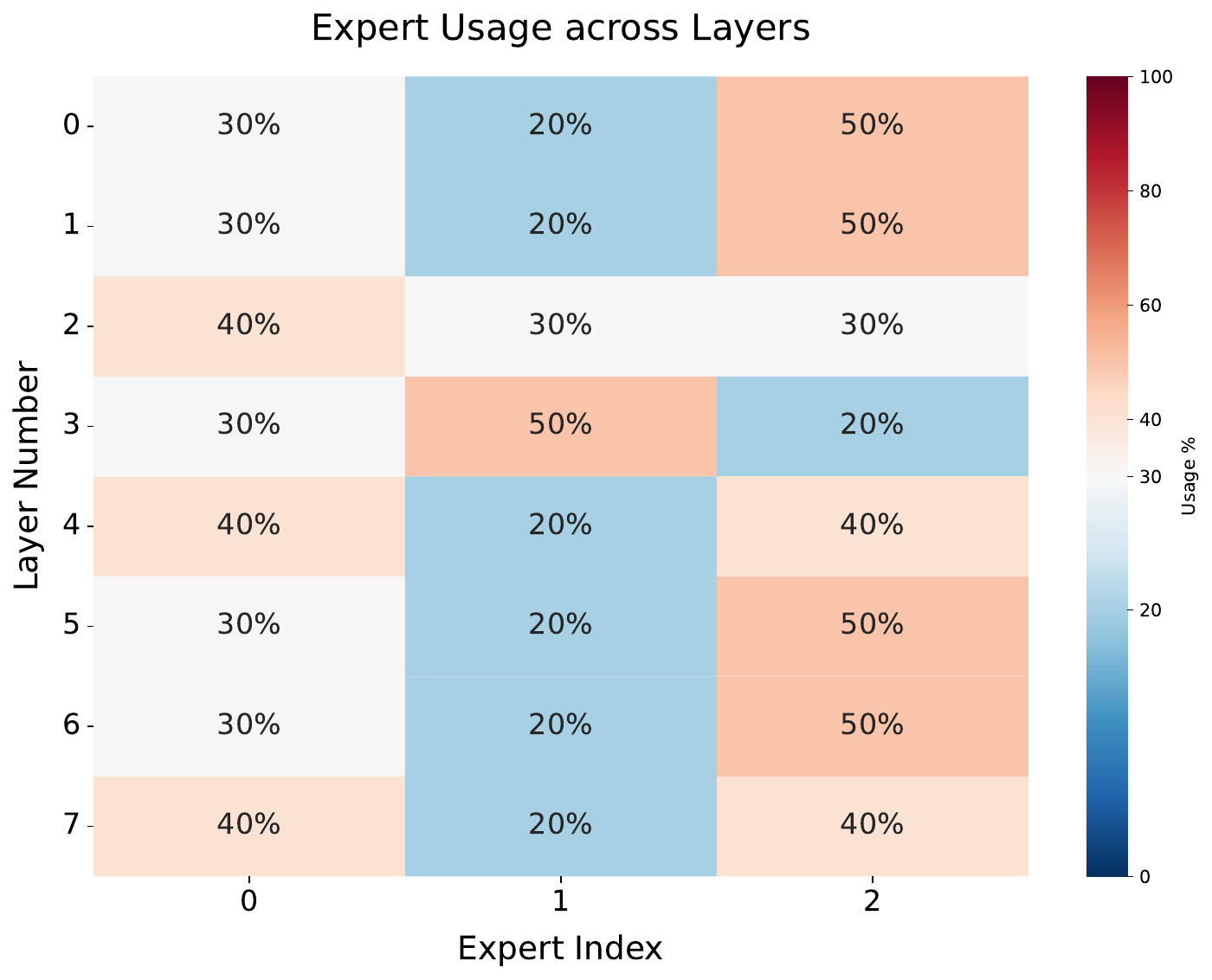} 
        \label{fig_3_experts}
    }
    \subfloat[4 Experts]{%
        \includegraphics[width=0.47\textwidth]{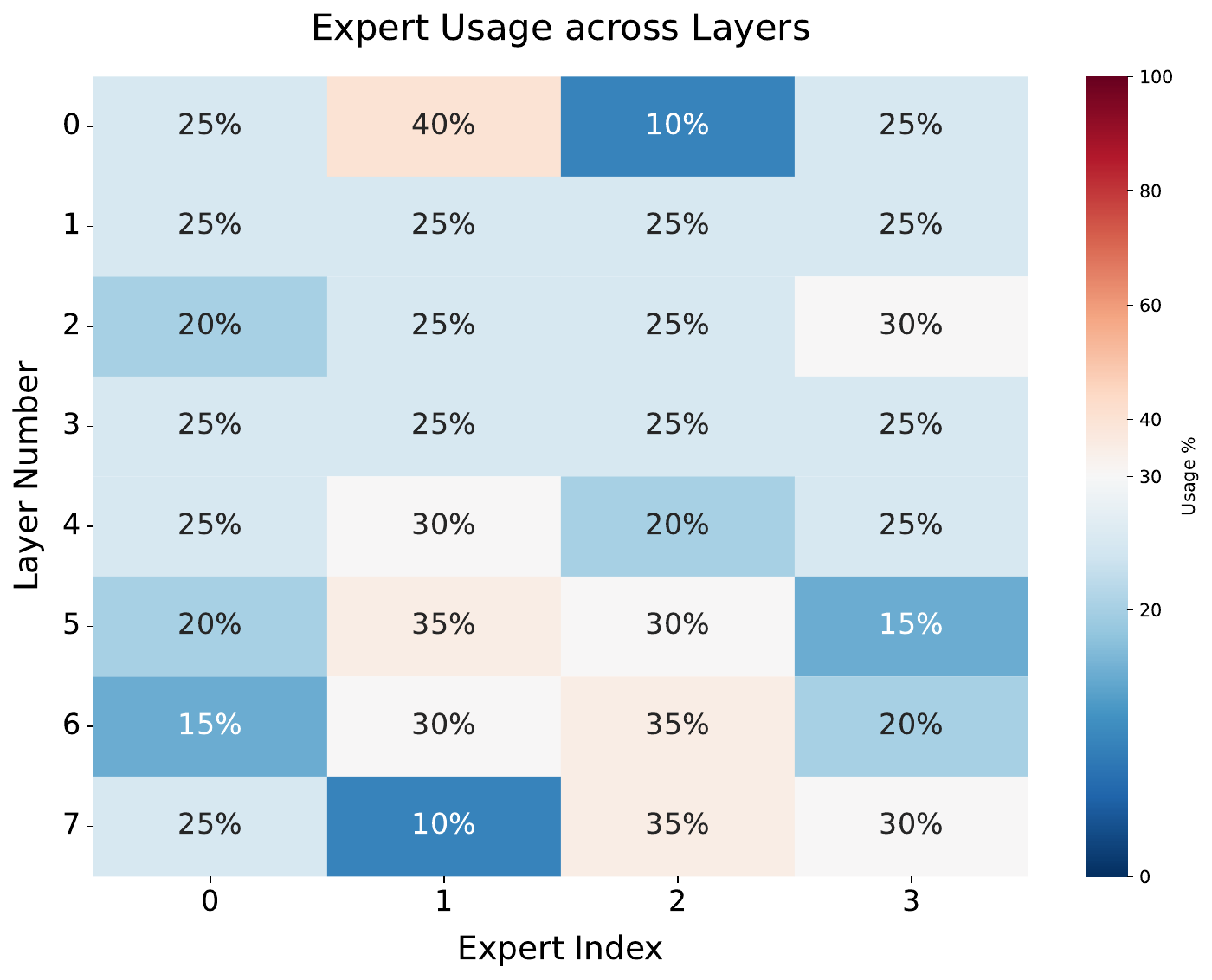} 
        \label{fig_4_experts}
    }
    \vspace{1em} 
    \subfloat[6 Experts]{%
        \includegraphics[width=0.47\textwidth]{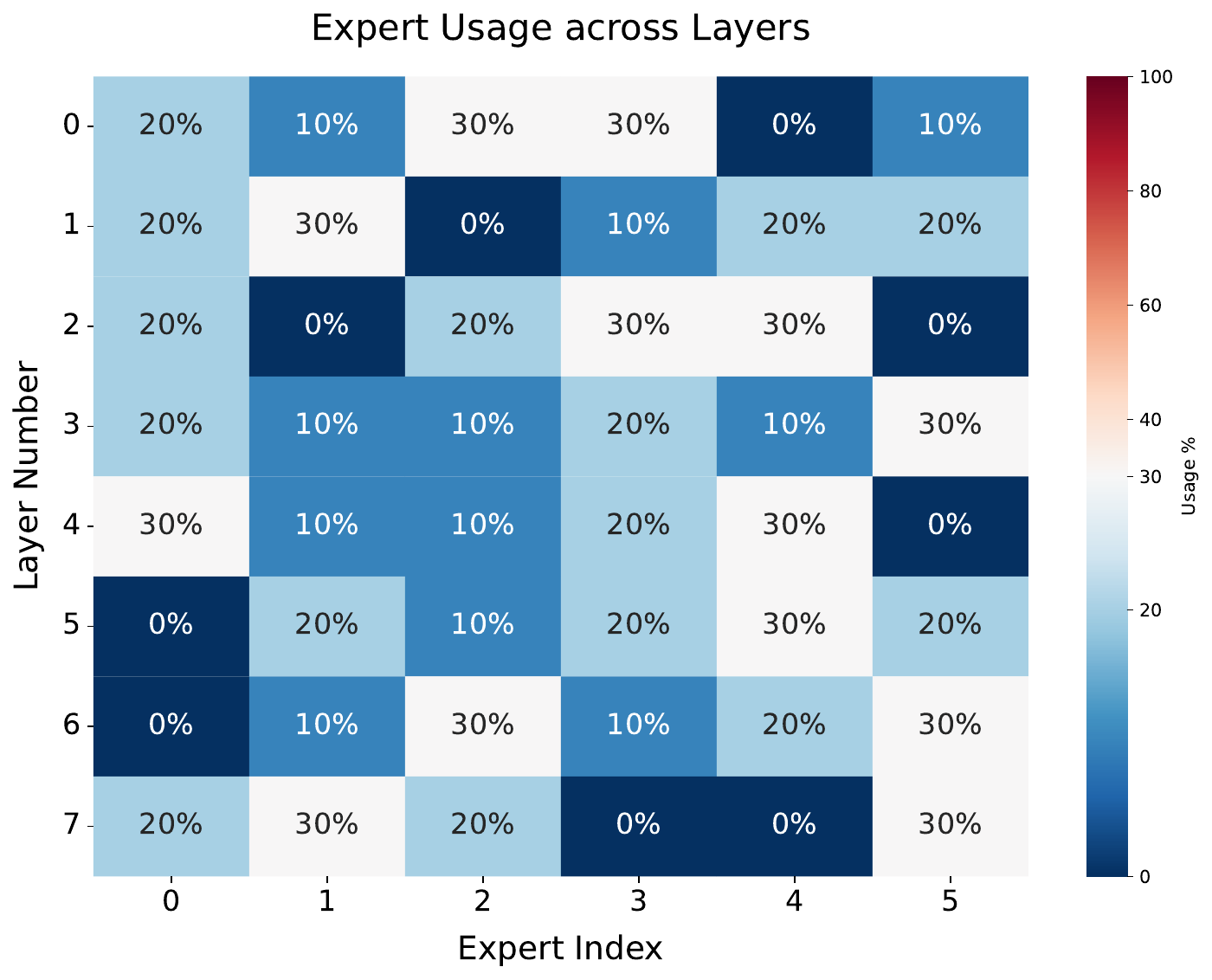} 
        \label{fig_6_experts}
    }
    \subfloat[8 Experts]{%
        \includegraphics[width=0.47\textwidth]{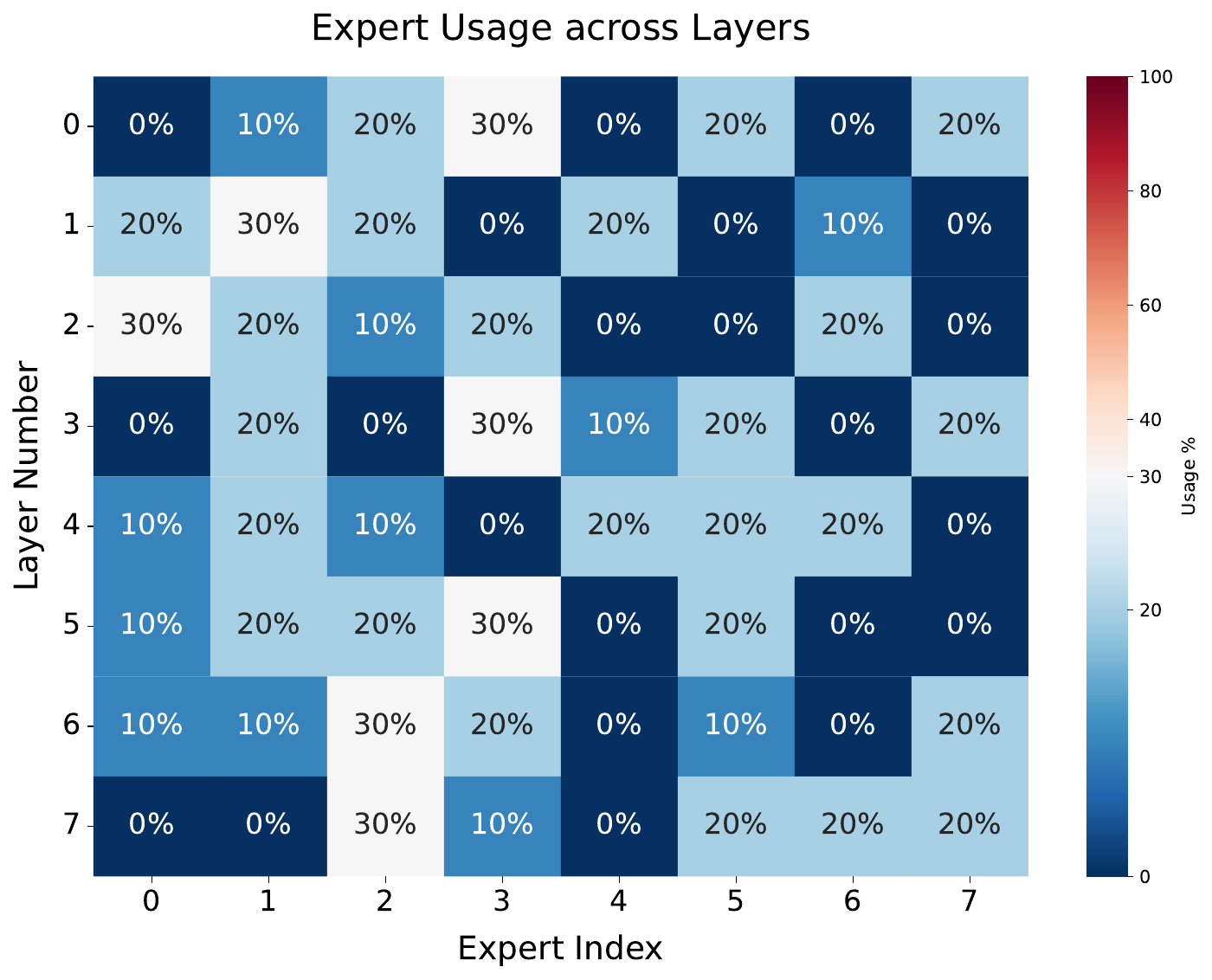}
        \label{fig_8_experst}
    }
    \caption{Average Expert Utilization for different number of used experts across all denoising levels.}
    \label{fig:Expert_Distribution_n_experts}
\end{figure}

We analyze how the load balancing loss affects expert distribution across noise levels by training \gls{mode} on LIBERO-10 with varying load balancing weights  $\gamma_{LB} \in [0.1, 0.01, 0.001, 0.0001 ]$
\autoref{fig:Expert_Distribution LB} visualizes the resulting expert distributions.

With a high load balancing loss of $0.1$, experts are used almost uniformly across all layers with minor variations in two out  of eight layers (\autoref{fig:lb0.1}). 
However, this enforced uniformity comes at a cost - the average performance drops to 0.9.
This result suggests that enforcing equal expert usage across noise levels may constrain the model's learning capacity.

At $0.01$, we observe a more flexible distribution while maintaining good overall expert utilization (\autoref{fig:lb0.01}). 
Within individual layers, expert usage varies from $10\%$ to $40\%$ in various layers.
This enables for specialization while preventing any expert from becoming dominant.
This configuration achieves the best average performance and provides evidence that moderate specialization benefits the policy most.

As we reduce $\gamma_{LB}$ further to a range of $[0.001, 0.0001]$, the expert distribution gets worse and the model there is an expert collapse is happening for lb of 0.0001 in several layers (\autoref{fig:lb0.001} and \autoref{fig:lb0.0001}).
These values show, that a the load balancing weight is crucial to guarantee a good usage of all experts for the model. 
The lower performance of these models (0.86 for $\gamma_{LB}=0.001$ and 0.83 for $\gamma_{LB}=0.0001$) suggests that balanced expert participation is important for optimal Diffusion Policy performance.

\textbf{Q: What happens with increased number of Experts?}
Next, we study the load balancing distribution for models that use more experts. 
Our experiments with 6 and 8 expert variants on LIBERO-10 reveal challenges in scaling beyond 4 experts. As shown in \autoref{fig:Expert_Distribution_n_experts}, configurations with more experts consistently underutilize their capacity, with most layers actively using only 4 experts even with increased load balancing regularization ($\gamma=0.1$). 
These larger configurations also show decreased performance, indicating that 4 experts provides an optimal balance between representation capacity and learning efficiency. 
We hypothesize that additional experts impair the model's ability to learn effective representations from image and language inputs while maintaining noise-level specialization. 
This phenomenon persists even when increasing the load balancing loss factor, further supporting our finding that 4 experts offers the best trade-off between model capacity and performance.

\begin{figure}[htb]
    \centering
    \subfloat[topk=1]{%
        \includegraphics[width=0.47\textwidth]{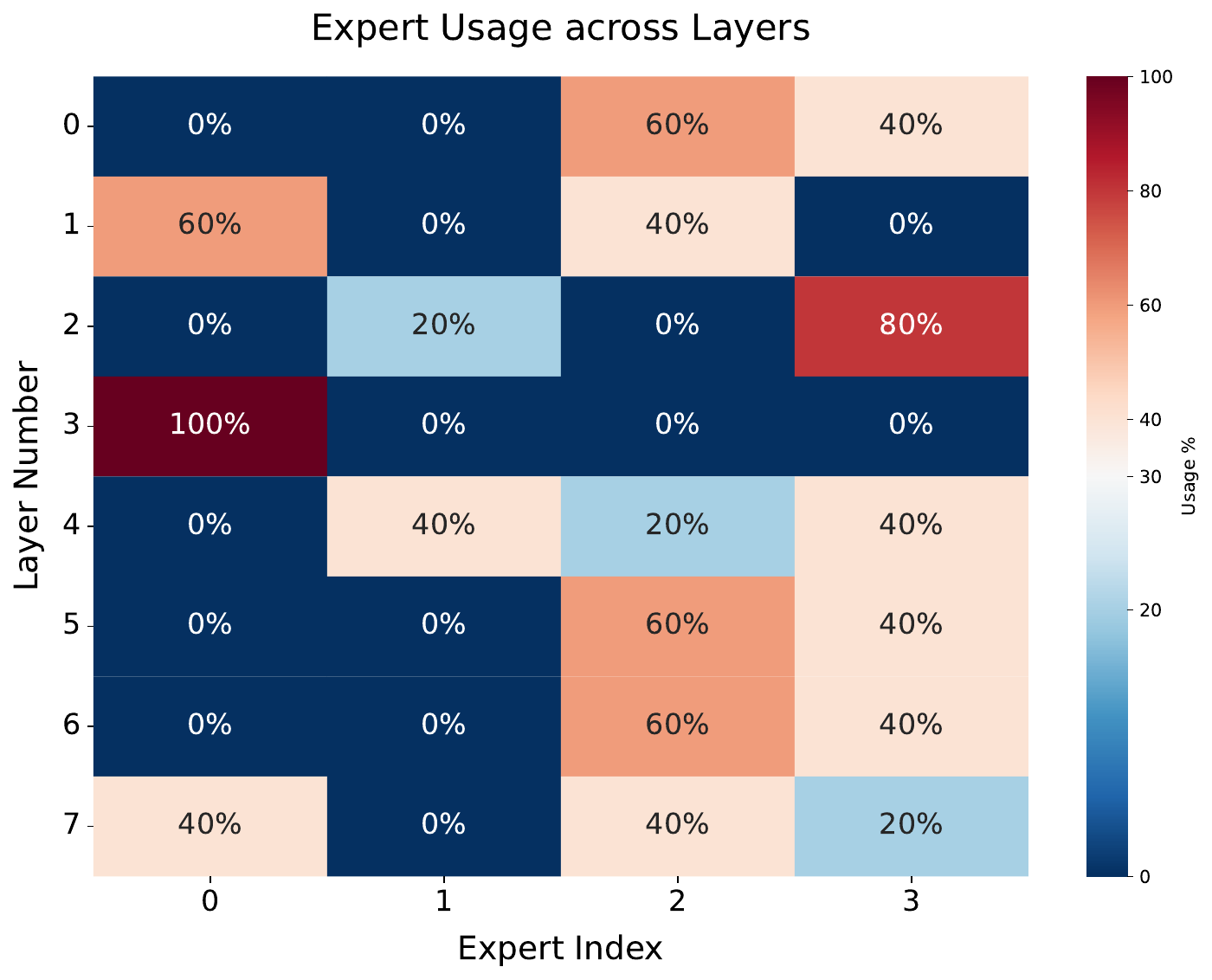}
        \label{fig:topk=1}
    }
    \subfloat[topk=1 with shared Expert]{%
        \includegraphics[width=0.47\textwidth]{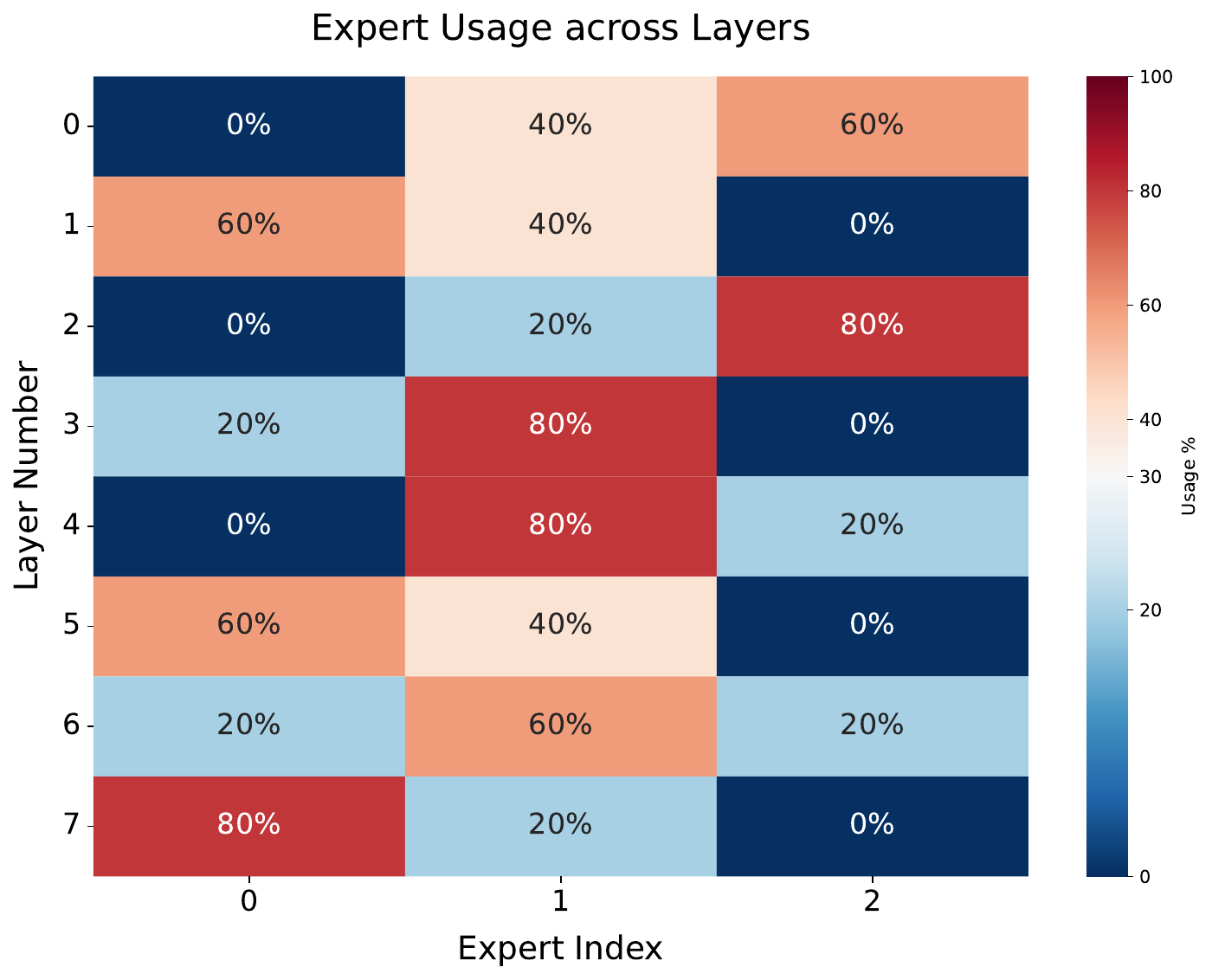} 
        \label{fig:topk=1_shared}
    }
    \caption{Average Expert Utilization for two variants that use topk=1: left MoDE with 4 experts, right MoDE with 4 experts, where 1 is used in all settings as a shared one. Overall, MoDE struggles to equally distribute tokens across experts in this setting.}
    \label{fig:Expert_Distribution_topk=1}
\end{figure}

\textbf{Q: How is the expert distribution with topk=1?}
We further study the impact of using $top_k =1$ for \gls{mode}.
We analyze MoDE's behavior with topk=1 routing, both with and without a shared expert, using 4 experts total.
As visualized in \autoref{fig:Expert_Distribution_topk=1}, topk=1 configurations struggle to maintain balanced expert utilization, typically collapsing to using only two of the four available experts. 
While this configuration shows only modest performance degradation on LIBERO-10, its impact is more visible on the challenging CALVIN ABC benchmark, where the average rollout length drops from 3.34 (topk=2) to 2.72 (topk=1). 
These results highlight the importance of expert interpolation across noise levels, with topk=2 providing the necessary flexibility to blend expert specializations for optimal performance.

\textbf{Q: What expert distribution patterns emerge for different noise levels?}

\begin{figure*}[ht]
\hspace{-1em}
    \centering
    \includegraphics[width=\columnwidth]{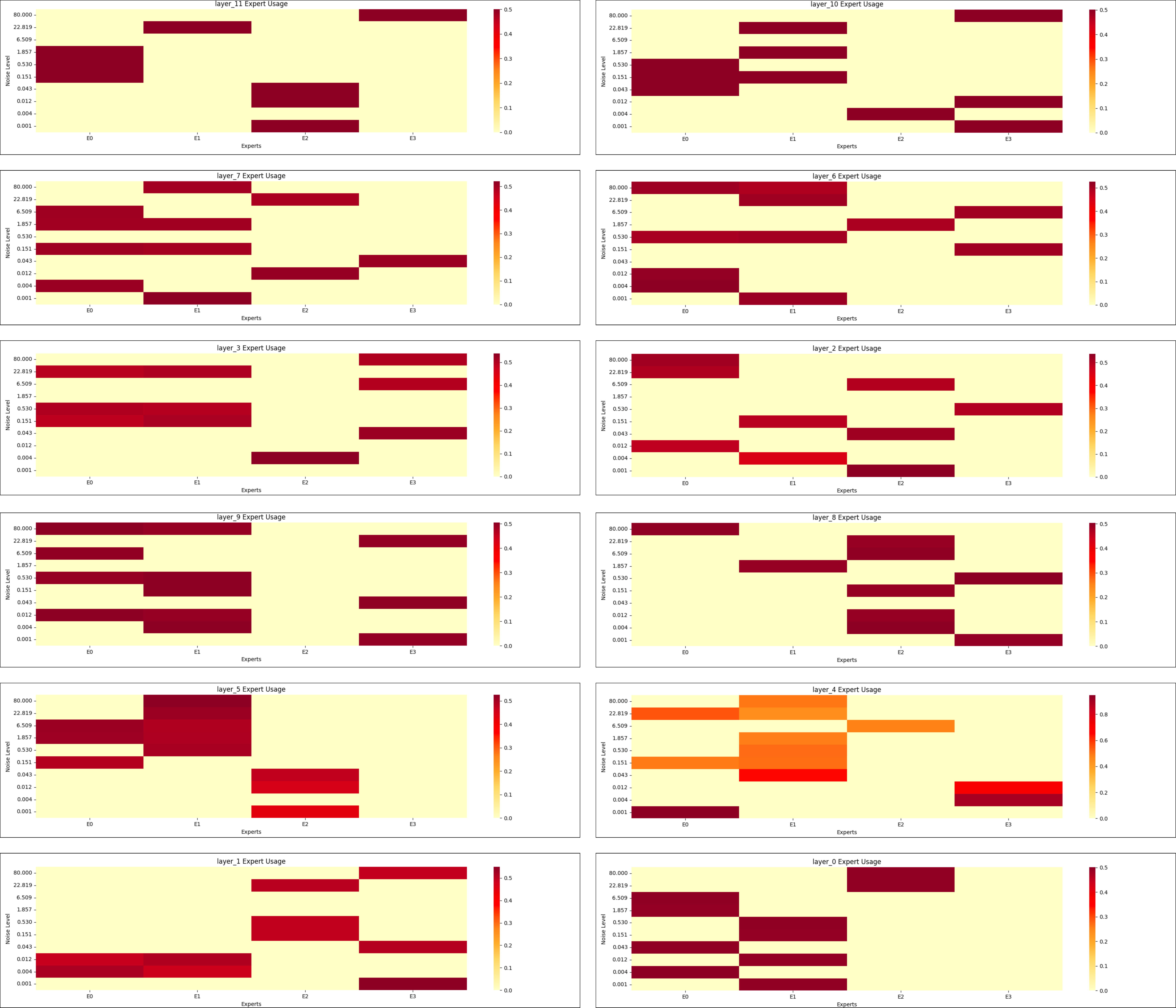}
    \caption{This image shows the expert usage distribution across 12 layers in a Mixture of Experts (MoE) model. Each subplot represents a different layer (from layer 0 to layer 11), with experts labeled E0, E1, E2, and E3 on the x-axis. The y-axis indicates the log-scaled token count, displaying the frequency of tokens routed to each expert within that layer. The color gradient indicates the proportion of tokens assigned to each expert, where darker colors represent higher usage.}
    \label{fig:pretrained_expert_distribution}
\end{figure*}

Finally, we analyze the different experts loads for various noise levels of our biggest model after pretraining on diverse robotics data. 
Detailed results are shown in \autoref{fig:pretrained_expert_distribution}.

Our analysis of the expert distribution reveals several key insights about how the model learns to organize its mixture-of-experts architecture:

\begin{itemize}
    \item \textbf{Noise-Level Specialization}
   \begin{itemize}
       \item Experts specialize between high-noise ($\sigma_1$-$\sigma_7$) and low-noise ($\sigma_8$-$\sigma_{10}$) regimes, particularly visible in L4 where E3 and E0 show distinct high activation patterns in different noise regimes
       \item Clear transition point around $\sigma_8$ where expert utilization patterns shift, most evident in L4 and L5 with stark changes in expert activation
       \item Smooth handoff between denoising phases, exemplified in L7 where expert activations gradually transition across noise levels
   \end{itemize}

   \item \textbf{Layer-wise Organization}
   \begin{itemize}
       \item Early layers (L0-L4) show distinct expert specialization, particularly visible in L4's strong alternating expert patterns
       \item Middle layers (L5-L7) demonstrate more distributed expert utilization, shown by more balanced activation patterns
       \item Later layers (L8-L11) return to clearer specialization, evident in L11's distinct expert preferences
       \item First layer (L0) shows particularly strong specialization for low-noise scenarios, with clear expert preferences in the final denoising steps
   \end{itemize}

   \item \textbf{Expert Role Distribution}
   \begin{itemize}
       \item Different experts develop specialized roles, most clearly visible in L4 where experts show strong preferences for specific noise ranges
       \item Some experts consistently handle high-noise denoising (evident in L3 and L4) while others focus on low-noise refinement (for example in in L7-L8)
       \item "Transitional experts" emerge in middle noise ranges, particularly visible in L5's balanced activation patterns
       \item Model employs different expert combinations across layers, well demonstrated in the contrasting patterns between L4 and L8
   \end{itemize}

   \item \textbf{Load Balancing Properties}
   \begin{itemize}
       \item Effective load balancing achieved across experts, particularly visible in the distributed patterns of L5-L7
       \item No single expert dominates across all noise levels, as shown by the varied activation patterns in each layer
       \item Smooth transitions between noise regimes, most evident in the gradual activation changes in L4 and L5
   \end{itemize}
\end{itemize}

These findings provide strong evidence that \gls{mode} effectively partitions the denoising process across its experts, with each expert naturally specializing in different phases. The emergence of this organization without explicit supervision supports the effectiveness of our architectural choices and routing strategy.

\subsection{Extended Related Work}

\textbf{MoE in Robotics.}
In the context of robotics, \gls{moe} models are used in many settings without being combined with a transformer architecture. 
Several works use a mixture of small MLP policies, that focus on different skills in Reinforcement Learning \citep{obando2024mixtures, pmlr-v164-celik22a, celik2024acquiring} or for robot motion generation\citep{hansel2023hierarchical, le2023hierarchical}, another body of work utilizes combinations of small CNNs robot perception \citep{riquelme2021scaling, mees2016choosing}.
Further applications include learning multimodal behavior using a mixture of Gaussian policies~\citep{blessing2023information, 10160543}.
Despite the extensive usage of \gls{moe} in many domains, no prior work has tried to utilize \gls{moe} together with Diffusion Policies for scalable and more efficient Diffusion Policies.

\textbf{Transformers for Robot Learning.}
Transformer models have become the standard network architecture for many end-to-end robot learning policies in the last few years.
They have been combined with different policy representations in the context of \gls{gcil}.
One area of research focuses on generating sequences of actions with Variational Autoencoder (VAE) models~\citep{bharadhwaj2023roboagent, zhao2023learning}. 
These action-chunking transformer models typically use an encoder-decoder transformer as a policy architecture. 
Several Diffusion Policies, such as Octo~\citep{octo_2023}, BESO~\citep{reuss2023goal}, ChainedDiffuser~\citep{xian2023chaineddiffuser} and 3D-Diffusion-Actor leverage a transformer model as a policy backbone.
Another direction of research treats behavior generation as discrete next-token predictions similar to auto-regressive language generation~\citep{touvron2023llama}.
C-Bet, RT-1, and RT-2 use discretized action binning to divide seen actions into $k$-classes~\citep{cui2023from, shafiullah2022behavior, rt12022arxiv, zitkovich2023rt}, while VQ-BeT~\citep{lee2024behavior} learns latent actions with residual Vector Quantization. 
Several works have shown the advantages of using pre-trained \gls{llm} or \gls{vlm} as a policy backbone, which are then finetuned for action generation~\citep{brohan2023rt, gu2023rttrajectory, open_x_embodiment_rt_x_2023, li2023vision}.
None of the recent work considers using any Mixture-of-Expert architecture for policy learning. 
\gls{mode} is the first architecture to leverage \gls{moe} architecture combined with diffusion for behavior generation.

\end{document}